\documentclass{article}
\usepackage[final]{neurips_2021}
\usepackage[utf8]{inputenc} 
\usepackage[T1]{fontenc}    
\usepackage{hyperref}       
\usepackage{url}            
\usepackage{booktabs}       
\usepackage{amsfonts}       
\usepackage{nicefrac}       
\usepackage{microtype}      
\usepackage{xcolor}         

\usepackage{todonotes}
\usepackage[normalem]{ulem}

\usepackage[acronym]{glossaries}
\usepackage{amsmath,amsthm}
\usepackage{amssymb}
\usepackage[ruled,vlined]{algorithm2e}

\usepackage{multirow}
\usepackage{bbm}

\usepackage[font=small,labelfont=bf]{caption}
\usepackage{subcaption}

\setcitestyle{numbers}

\definecolor{darkspringgreen}{rgb}{0.09, 0.45, 0.27}

\newtheorem*{assumption*}{\assumptionnumber}
\providecommand{\assumptionnumber}{}
\makeatletter
\newenvironment{assumption}[1]
 {%
  \renewcommand{\assumptionnumber}{Assumption #1}%
  \begin{assumption*}%
  \protected@edef\@currentlabel{#1}%
 }
 {%
  \end{assumption*}
 }
\makeatother

\newtheorem{theorem}{Theorem}
\newtheorem{lemma}{Lemma}
\newtheorem{definition}{Definition}

\makeglossaries
\newacronym{DSDI}{mDSDI}{meta-Domain Specific-Domain Invariant}
\newacronym{DG}{DG}{Domain Generalization}

\DeclareMathOperator{\Cov}{Cov}

\title{Exploiting Domain-Specific Features to Enhance Domain Generalization}

\author{%
	Manh-Ha Bui$^{1}$ \hspace{0.1cm}
	Toan Tran$^{1}$ \hspace{0.1cm}
	Anh Tuan Tran$^{1}$ \hspace{0.1cm} 
	Dinh Phung$^{1,2}$ \\
	$^{1}$ VinAI Research, Vietnam \hspace{0.1cm} 
	$^{2}$ Monash University, Australia \\
	\texttt{\{v.habm1, v.toantm3, v.anhtt152, v.dinhpq2\}@vinai.io}
   
  \thanks{Correspondence to Manh-Ha Bui: <\texttt{hb.buimanhha@gmail.com}>.}
}

\begin{document}

\maketitle

\begin{abstract}
\acrfull{DG} aims to train a model, from multiple observed source domains, in order to perform well on unseen target domains. To obtain the generalization capability, prior \acrshort{DG} approaches have focused on extracting domain-invariant information across sources to generalize on target domains, while useful domain-specific information which strongly correlates with labels in individual domains and the generalization to target domains is usually ignored. In this paper, we propose \acrfull{DSDI} - a novel theoretically sound framework that extends beyond the invariance view to further capture the usefulness of domain-specific information. Our key insight is to disentangle features in the latent space while jointly learning both domain-invariant and domain-specific features in a unified framework. The domain-specific representation is optimized through the meta-learning framework to adapt from source domains, targeting a robust generalization on unseen domains. We empirically show that \acrshort{DSDI} provides competitive results with state-of-the-art techniques in \acrshort{DG}. A further ablation study with our generated dataset, Background-Colored-MNIST, confirms the hypothesis that domain-specific is essential, leading to better results when compared with only using domain-invariant.
\end{abstract}
\section{Introduction and Related work}
\acrfull{DG} has recently become an important research topic in machine learning due to its real-world applicability and its close connection to the way humans generalize to learn in a new domain. In a \acrshort{DG} framework, the learner is trained on multiple datasets collected under different environments without any access to any data on the target domain~\cite{blanchard2011generalizing}. One of the most notable approaches to this problem is to learn the ``domain-invariant'' features across these training datasets, with the assumption that these invariant representations are also held in unseen target domains~\cite{li2017deeper,Li2018DeepDG,hu2019domain,ilse2019diva,chuang2020estimating}. While this has been shown to work well in practice, its key drawback is completely ignoring ``domain-specific'' information that could aid the generalization performance, especially when the number of source domains increases~\cite{chattopadhyay2020learning}. \\
For instance, consider the problem of classifying dog or fish images from two source domains: sketch and photo. While the sketch contains a conceptual drawing of the animal, the photo includes their taken picture within a background. In this case, sketch domain-invariant is kept across domains, while domain-specific, e.g., a dog in a house or fish in the ocean, will be discarded due to only existing in the photo domain. However, this background information, when present, could lead to an improvement of the classification performance in target domains due to common association between the objects of its background, and when negligent sketches are hard to distinguish. From a theoretical standpoint, there has also been strong recent evidence to indicate the insufficiency of learning domain-invariant representation for successful adaptation in domain adaptation problems~\cite{zhao19onlearning,johansson2019support}. For example, Zhao et al.~\cite{zhao19onlearning} has pointed out the degradation in target predictive performance if domain-invariant representations are forced while the marginal label distributions on the source and target domains are overly different.

Utilizing domain-specific features in \acrshort{DG} has been widely studied in recent works (e.g.,~\cite{ding17lowrank,chattopadhyay2020learning}). Ding and Fu~\cite{ding17lowrank} introduce multiple domain-specific networks for each domain, then use the structured low-rank constraints to align them with domain-invariant. While this encourages the better transfer of knowledge, its main problem is the requirement of too many domain-specific networks. More recently, Chattopadhyay et al.~\cite{chattopadhyay2020learning} proposed a masking strategy to disentangle domain-invariant and domain-specific to further boost domain-specific learning, but its key drawback is that domain-invariant/domain-specific representations might not be disentangled since the learning and inferring procedures are performed implicitly (i.e., without any theoretical guarantee) through a mask generalization process. That means it lacks a clear motivation as well as theoretical justifications.

Regarding meta-learning related work, a typical approach involving meta-learning in \acrshort{DG} is MLDG~\cite{li2017learning} that is based on gradient update which simulates train/test domain shift within each mini-batch, mainly to learn transferable weight representations from meta-source domains to quickly adapt to the meta-target domain, and so improve generalization ability. However, their task objective adapts for all representation features which include domain-invariant, since low effectiveness because domain-invariant is stable across domains, pushing to adapt those features might affect the stability of those domain-invariant, leading to a lower generalization performance on the target domain.

To handle these domain-invariant shortcomings, in this paper, we propose a novel theoretically sound \acrshort{DG} approach that aims to extract label-informative domain-specific and then explicitly disentangles the domain-invariant and domain-specific representations in an efficient way without training multiple networks for domain-specific. Following the meta-learning idea and mitigating previous work's drawbacks, we apply a meta-learning technique specifically to exploit domain-specific quality which should need to be adapted to unseen domains from source domains. Our contributions in this work are summarized as follows:
\begin{itemize}
    \item We provide a theoretical analysis based on the information bottleneck principle to point out the limitation of only learning invariant and the importance of domain-specific representation by a certainly plausible assumption.
    \item We then develop a rigorous framework to formulate elements of domain-invariant/domain-specific representations, in which our key insight is to introduce an effective meta-optimization training framework~\cite{li2017learning} to learn domain-specific representation from multiple training domains. Without accessing any data from unseen target domains, the meta-training procedure provides a suitable mechanism to self-learn domain-specific representation. We term our approach \acrfull{DSDI} and provide necessary theoretical verifications for it.
    \item To demonstrate the merit of the proposed \acrshort{DSDI} framework, we extensively evaluate \acrshort{DSDI} on several state-of-the-art \acrshort{DG} benchmark datasets, including Colored-MNIST, Rotated-MNIST, VLCS, PACS, Office-Home, Terra Incognita, DomainNet in addition to our newly created Background-Colored-MNIST for the ablation study to examine the behavior of our \acrshort{DSDI}.
\end{itemize}
\section{Methodology}
\subsection{Problem setting and Definitions}
Let $\mathcal{X} \subset \mathbb R^D$ be the sample space and $\mathcal{Y} \subset \mathbb R$ the label space. Denote the set of joint probability distributions on $\mathcal{X} \times \mathcal{Y}$ by $\mathcal{P}_{\mathcal{X} \times \mathcal{Y}}$, and the set of probability marginal distributions on $\mathcal{X}$ by $\mathcal{P}_{\mathcal{X}}$. A domain is defined by a joint distribution $P(x, y) \in \mathcal{P}_{\mathcal{X}\times \mathcal{Y}}$, and let $\mathcal P$ be a measure on $\mathcal{P}_{\mathcal{X} \times \mathcal{Y}}$, i.e., whose realizations are distributions on $\mathcal{X} \times \mathcal{Y}$.

Denote $N$ source domains by $S^{(i)} = \{(x_{j}^{(i)}, y_{j}^{(i)})\}_{j=1}^{n_i}, \; i=1,\ldots,N$, where $n_i$ is the number of data points in $S^{(i)}$, i.e., $(x_{j}^{(i)}, y_{j}^{(i)}) \overset{iid}{\sim} P^{(i)}(x,y)$ where $P^{(i)}(x,y) \sim \mathcal P$; and $x_j^{(i)} \sim P^{(i)}_{\mathcal X}$, in which $P^{(i)}_{\mathcal X} \sim P_{\mathcal X}$. In a typical DG framework, a learning model which is only trained on the set of source domains $\{S^{(i)}\}_{i=1}^N$ without any access to the (unlabeled) data points in the target domain, arrives at a good generalization performance on the test dataset $S^T = \{(x_{j}^T, y_{j}^T)\}_{j=1}^{n_T}$, where $(x_{j}^T, y_{j}^T) \overset{iid}{\sim} P^T(x,y)$ and $P^T(x,y) \sim \mathcal{P}$.

First, we present the definition of domain-invariant representation in a latent space $\mathcal Z$ under covariate shift assumption (i.e., the conditional distribution $P(y\vert x)$ is unchanged across the source domains):
\begin{definition}\label{def:dom_inv}
A feature extraction mapping $Q: \mathcal X \to \mathcal Z$ is said to be {\bf domain-invariant} if the distribution $P_Q(Q(X))$ is unchanged across the source domains, i.e.,
$\forall i,j=1,\ldots,N, \; i \neq j$ we have $P_Q^{(i)}(Q(X)) \equiv  P_Q^{(j)}(Q(X))$, where $P_Q^{(i)}(Q(X))=P_Q(Q(X)\vert X \sim P^{(i)}_{\mathcal X}), \; i=1,\ldots,N$. In this case, the corresponding latent representation $Z_I = Q(X)$ is then called the domain-invariant representation (see~\cite{Li2018DeepDG} also).   
\end{definition} 

As mentioned in the example in the introduction part, the definition~\ref{def:dom_inv} reveals that the extracted domain-invariant latent $Z_I$ could be the conceptual drawing of the animal which is shared in both sketch and photo domains. However, when existing background information is taken by a picture such as a house or ocean, it is crucial to take these backgrounds into account because the domain-invariant feature extraction $Q$ might ignore them by only existing in the photo domain. Therefore, we next introduce the definition of domain-specific in latent space as follows:
\begin{definition}\label{def:dom_spe}
A feature extraction mapping $R: \mathcal X \to \mathcal Z$ is said to be {\bf domain-specific} if $\; \forall i,j=1,\ldots,N, \; i \neq j$ such that $P_R^{(i)}(R(X)) \neq  P_R^{(j)}(R(X))$, where $P_R^{(i)}(R(X))=P_R(R(X)\vert X \sim P^{(i)}_{\mathcal X}), \; i=1,\ldots,N$. In this case, given $X \sim P^{(i)}_{\mathcal X}$ the corresponding latent representation $Z^{(i)}_S = R(X)$ is then called the domain-specific representation w.r.t. the domain $S^{(i)}$.
\end{definition}

Definition~\ref{def:dom_spe} states that for any domain $i$ and $j$, the distributions of the domain-specific latent $P_R^{(i)}(R(X))$ and $P_R^{(j)}(R(X))$ must be completely different. For instance, following our mentioned example, given two fish images from $i$ and $j$ that are sketch and photo domain, the mapping $R(X)$ should extract specific information that only belongs to the domain including the shadow of the fish drawing in the sketch and ocean background information in the photo domain. 

To this end, this paper aims to show that only learning domain-invariant will limit the prediction performance and generalization ability. Hence, we next provide a formal explanation for the motivation of learning domain-specific, by showing the potential drawback of only learning domain-invariance in terms of predicting class labels.

\subsection{A theoretical analysis under the Information bottleneck method}
\begin{figure}[ht!]
\vskip -0.1in
\begin{center}
  \includegraphics[width=1.0\linewidth]{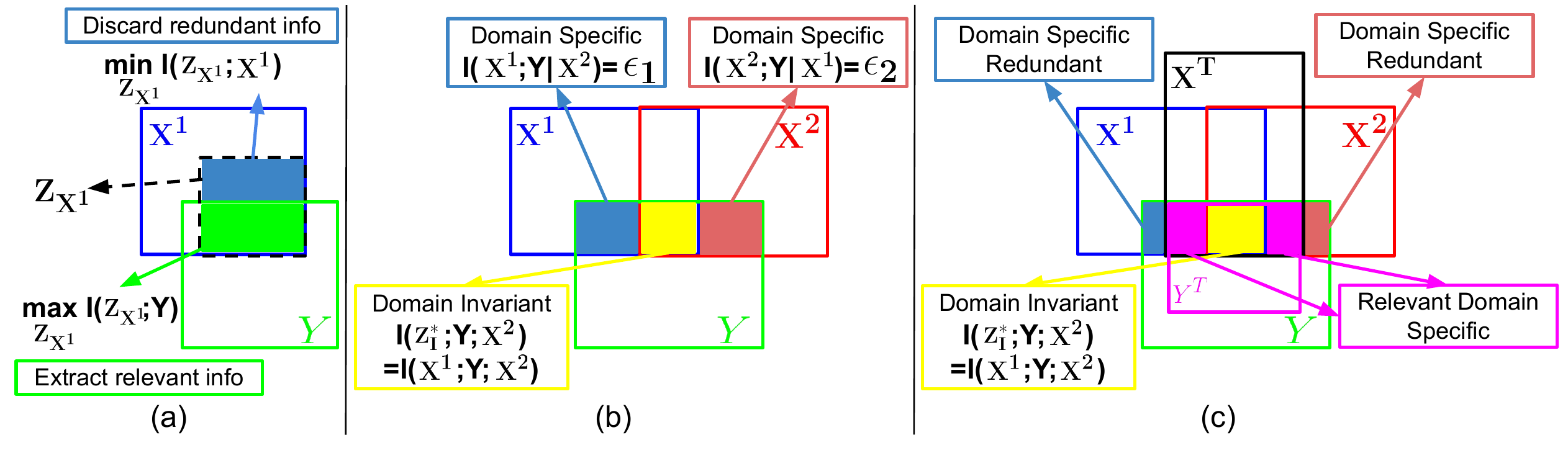}
\end{center}
  \caption{Venn diagram showing relationships between source domains represented by $X^1$, $X^2$, target domain represented by $X^T$, and label $Y$. (a) The learning procedure of minimal and sufficient label-related representation in definition~\ref{def:min_suf}. (b) Explaining the theorem~\ref{theo:inv} where the domain-invariant based method provides an inferior prediction performance to our proposed method that incorporates both domain-invariant $I(Z_{I^*};Y)$ and label-related domain-specific values $\epsilon$ made by our assumption~\ref{asm:specific}. (c) A case when the unseen (target) domain has different $X^T$ and $Y^T$, while domain-invariant information is still stable across domains, some domain-specific in source domains become redundant information.}
\label{fig:insights}
\vskip -0.1in
\end{figure}

\textbf{Notations.} 
Given three arbitrary random variables $A,B$, and $C$, let us use $I(A;B)$ to represent mutual information between $A$ and $B$; $I(A;B|C)$ to represent conditional mutual information of $A$ and $B$ given $C$; $H(A)$ to represent entropy of $A$; and $H(A|B)$ to represent conditional entropy for random variables $A$ given $B$. For simplicity, we consider the case with two source domains $S^1,S^2$ (the results with multiple source domains can be naturally extended from there). We also define two corresponding random variables $X^{i} \sim P^{(i)}_{\mathcal X}$, that are sampled from the marginal distribution $P_{\mathcal X}$ in the domain $S^{i}, \; i=1,2$.

Figure~\ref{fig:insights} illustrates all the definitions and assumptions above used for our theoretical verification (in Theorem~\ref{theo:inv}). In particular, in that figure, each of the four colored rectangles represents an individual entropy: $H(X^1)$ for domain $S^1$ is in blue, $H(X^2)$ for domain $S^2$ is in red, $H(X^T)$ for the target domain is in black, and $H(Y)$ for the class label is in green border rectangle.

We first show the ineffectiveness of only learning domain-invariant information when compared with incorporating domain-specific in source domain $S^1$ (and similarly with domain $S^2$). Our justification partly relies on the following assumption about the correlation between the domain-specific representation and the class label:

\begin{assumption}{1} \label{asm:specific}
(Label-correlated domain-specificity) Assuming that
there exists a domain-specific representation $Z_S^{(1)}$ extracted by the deterministic mapping $Z_{S}^{(1)}=R(X^1)$ in definition~\ref{def:dom_spe}, which correlates with label in domain $S^1$ such that $I(Z_{S}^{(1)};Y|X^2)=I(X^1;Y|X^2)=\varepsilon_1$, 
where $\varepsilon_1>0$ is a constant.
\end{assumption}

Assumption~\ref{asm:specific} indicates that, for the source domain $S^1$, we can learn $Z_{S}^{(1)}=R(X^1)$ such that $I(Z_{S}^{(1)};Y|X^2)$ is strictly positive and equals to $I(X^1;Y|X^2)$, where $I(X^1;Y|X^2)$ is the specific information that correlates with the label in the domain $S^1$, but not in the domain $S^2$~\cite{tishby99information}. For instance, in the example mentioned in the introduction, if domain $S^1$ is “photo” while $S^2$ is “sketch”, the value of $\epsilon_1$ should be positive because the background information such as a house, the ocean also provides information to predict whether the object is a dog or fish without considering its conceptual drawing. This assumption is particularly valid and practically plausible and is demonstrated by several examples observed in our experiments. For instance, for the DomainNet benchmark dataset, in the real-world domain, many bed pictures contain a bed in the room or bike pictures that have bicycles parked on the street. Other examples are in PACS such as dogs in the yard or guitars lying on a table in photo and art domains. These examples are strongly related to assumption~\ref{asm:specific}, in which specific information correlates with labels in a particular domain.

We next present supervised learning frameworks under the umbrella of the information theory~\cite{Tishby99theinformation,tsai2021selfsupervised} and the information bottleneck method~\cite{Tishby99theinformation,achille2018emergence} that generalizes minimal sufficient statistics to the minimal (i.e., less complexity) and sufficient (i.e, better fidelity) representations. The learning process of such representations is equivalent to solving the following objectives:

\begin{definition}\label{def:min_suf}
(Minimal and sufficient representations with label~\cite{tsai2021selfsupervised}). Let $Z_{X^1}=G(X^1)$ is the output of a deterministic latent mapping $G$. A representation $Z_{sup}$ is said to be the sufficient label-related representation and $Z_{\sup^*}$ is said to be the minimal and sufficient representation if:
\begin{align*}Z_{sup} = \underset{G}{\mathrm{argmax}}I(Z_{X^1};Y) \text{ and } Z_{\sup^*} = \underset{Z_{\sup}}{\mathrm{argmin}} I(Z_{\sup};X^1) \text{ s.t. } I(Z_{\sup};Y) \text{ is maximized.}
\end{align*}
\end{definition}

The learning procedure for definition~\ref{def:min_suf} is illustrated in Figure~\ref{fig:insights}: (a). The method is equivalent to employ compressed representations to reduce the complexity (redundant information) of $I(Z_{X^1};X^1)$ by minimizing and providing sufficient representation to class label $Y$ by maximizing $I(Z_{X^1};Y)$. Similarly and motivated by multi-view information bottleneck settings~\cite{federici2020multiview}, we present the objective of learning sufficient (and minimal) representations with domain-invariant information in the below definition:

\begin{definition}\label{def:min_inv}
(Minimal and sufficient representations with domain-invariance~\cite{federici2020multiview}). Let $Z_{X^1}=Q(X^1)$ is the output of a deterministic domain invariant mapping $Q$ in the definition~\ref{def:dom_inv}. Then $Z_{I}$ is said to be the sufficient domain-invariant representation and $Z_{I^*}$ is said to be the minimal and sufficient representation if: 
\begin{align*}
Z_{I} = \underset{Q}{\mathrm{argmax}}I(Z_{X^1};X^2) \text{ and } Z_{I^*} &= \underset{Z_{I}}{\mathrm{argmin}} I(Z_{I};X^1) \text{ s.t. } I(Z_{I};X^2) \text{ is maximized.}
\end{align*}
\end{definition}

Definition~\ref{def:min_inv} introduces a learning strategy for domain-invariance across domains (or views) that preserves shared information across two domains by maximizing $I(Z_{X^1};X^2)$; and also reduces specificity (redundant information) of the domain $S^1$ by minimizing $I(Z_{X^1};X^1)$. We next present a lemma about the conditional independence between the latent representation $Z_{X^1}$ and both the label $Y$ and the random variable $X^2$ when $Q$, $R$, and $G$ are deterministic functions of the random variable $X^1$:

\begin{lemma}\label{lem:determinism}
(Determinism~\cite{tsai2021selfsupervised}) If $P(Z_{X^1}|X^1)$ is a Dirac delta function, then the following conditional independence holds: $Y\perp \!\!\! \perp Z_{X^1}|X^1$ and $X^2\perp \!\!\! \perp Z_{X^1}|X^1$, inducing a Markov chain $X^2 \leftrightarrow Y \leftrightarrow X^1 \rightarrow Z_{X^1}$.
\end{lemma}
The proof of Lemma~\ref{lem:determinism} is provided in Appendix~\ref{proof:lemma}.

Lemma~\ref{lem:determinism} simply states that $Z_{X^1}$ contains no more information than $X^1$. Now, we show the ineffectiveness of only learning domain-invariant approach, based on the existence of the label-related domain-specific in the following theorem:

\begin{theorem}\label{theo:inv} 
(Label-related information with domain-specificity) Assuming that there exists a domain-specific value $\varepsilon_1>0$ in domain $S^1$ (see Assumption~\ref{asm:specific}), the label-related representation - based learning approach (i.e., using $Z_{sup}$ and $Z_{sup^*}$) provides better prediction performance than the domain-invariant representation - based method (i.e., using $Z_{I}$ and $Z_{I^*}$). Formally,
\begin{align*}
I(X^1;Y)=I(Z_{sup};Y)=I(Z_{sup^*};Y)=I(Z_{I^*};Y)+\varepsilon_1>I(Z_{I^*};Y).
\end{align*}
\end{theorem}
The proof of Theorem~\ref{theo:inv} is mainly based on the result of Lemma~\ref{lem:determinism}, and is provided in Appendix~\ref{proof:theorem}.

The visualization of Theorem~\ref{theo:inv} is depicted by Figure~\ref{fig:insights}: (b), where the domain-specific value for domain $S^2$, $\varepsilon_2$ is obtained in the same way as $\varepsilon_1$. It indicates that if an existing domain-specific representation has the positive corresponding information value $\varepsilon$, the domain-invariance-based learning method provides an inferior prediction performance to our proposed method that incorporates both domain-invariant and label-related domain-specific.

Now, Theorem~\ref{theo:inv} suggests that besides optimizing a domain-invariant mapping $Q$ as usual, we should jointly optimize domain-specific mapping $R$ to achieve a better generalization performance. However, in domain generalization, we are not allowed to access the target domain for training and must use $Q$ and $R$ from source domains. As pointed out in~\cite{ding17lowrank,piratla2020efficient,chattopadhyay2020learning}, although domain-invariant might be the same because it is unchanged across source domains, there is no guarantee whether this domain-specific information on the source domain is relevant to the target domain while making the prediction. Figure~\ref{fig:insights}: (c) illustrates this case when the target domain has different $X^T$ and  $Y^T$, then some extracted domain-specific from source domains become redundant information. Therefore, the next raising question is how to learn domain-invariant and domain-specific effectively. 

To handle these shortcomings, we next propose a unified framework that jointly optimizes both $Q$ and $R$ by disentangling their feature representation. In particular, the deterministic mapping $Q$ is optimized by adversarial learning to extract useful domain-invariant features across domains. Meanwhile, by leveraging the transfer weight representations from meta-source domains to adapt to a meta-target domain, we apply meta-learning to deterministic mapping $R$ to force it to extract relevant domain-specific features of the target domain to improve generalization ability.

\subsection{Algorithm: \acrfull{DSDI}} 
So far, we have discussed the main ideas of our proposed method. Here we discuss implementation details for our proposed \acrshort{DSDI} approach.

\begin{figure}[ht!]
\vskip -0.1in
\begin{center}
\includegraphics[width=0.8\linewidth]{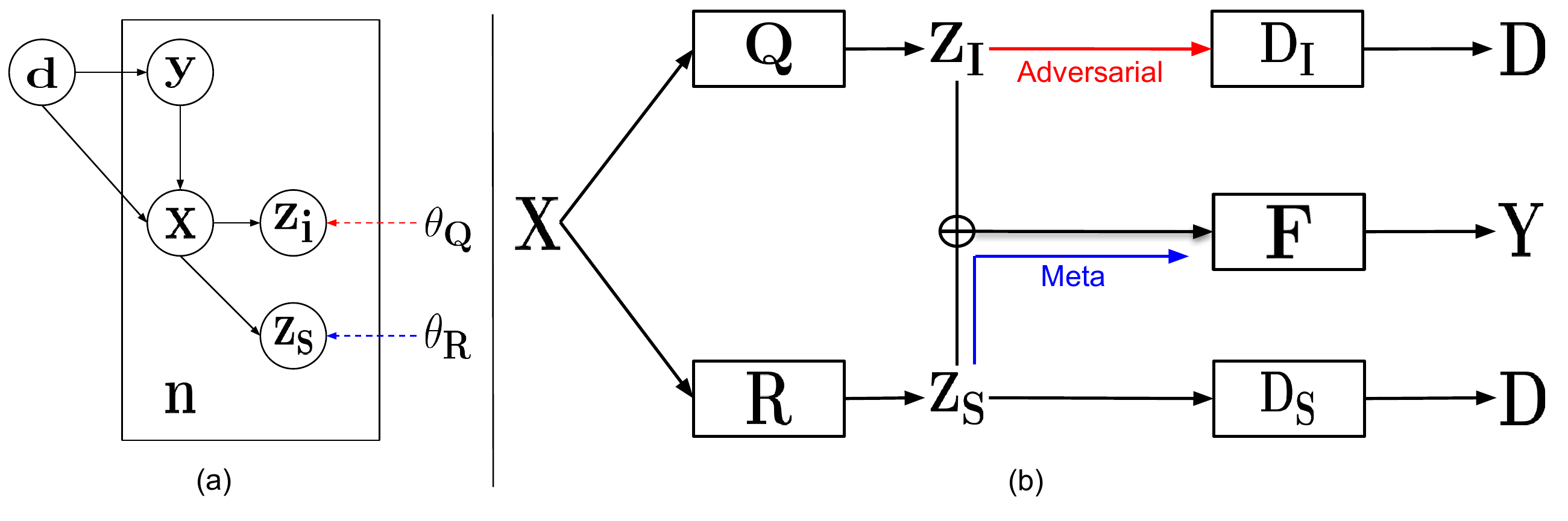}
\end{center}
   \caption{The graphical model (a) and overall architecture (b) for our proposed \acrshort{DSDI}, including: domain-invariant $Z_{I}$ is optimized via adversarial training with domain discriminator $D_{I}$, domain-specific $Z_{S}$ is optimized via domain classifier $D_S$, these latent $Z_I$ and $Z_S$ are disentangled by using covariance matrix. To push them to contain label information, these latents are integrated into a classifier $F$ which is optimized via cross-entropy with the label $Y$. To make the model able to adapt specific information from source to unseen domain while still remaining domain-invariance information across domains, we additionally push $Z_S$ through a meta-learning procedure.}
\label{fig:DSDI}
\vskip -0.2in
\end{figure}

Figure~\ref{fig:DSDI} shows the graphical model and overview of our \acrshort{DSDI} framework. In particular, our unified network consists of the following components: a domain-invariant representation $Z_{I}=Q_{\theta_{Q}}(X)$; a domain-specific representation $Z_{S}=R_{\theta_{R}}(X)$; a domain discriminator $D_{\theta_{D_I}}: Z_I \rightarrow \overline{1,N}$; a domain classifier $D_{\theta_{D_S}}: Z_S \rightarrow \overline{1,N}$ and a classifier $F_{\theta_F}: Z_I \oplus Z_S \to \mathcal Y$. We also denote domain random variable by $D$, sample space by $\mathcal D$, and outcome by $d$.

\textbf{Domain-Invariant and Domain-Specific Extraction.}
The domain-invariant representation $Z_I$ defined in Definition~\ref{def:dom_inv}, is obtained by using an adversarial training framework~\cite{Li2018DeepDG}, in which the domain discriminator $D_{\theta_{D_I}}$ tries to maximize the prediction probability of the domain label from the latent $Z_I$, while the goal of the encoder $Q_{\theta_Q}$ is to map the sample $X$ to the latent $Z_I$, such that $D_{\theta_{D_I}}$ cannot discriminate the domain of $X$. This task can be performed by solving the following min-max game:
\begin{equation}\label{eq:atloss}
    \underset{\theta_{Q}}{\min} \text{ } \underset{\theta_{D_{I}}}{\max}\left \{ L_{Z_{I}} := -\mathbb{E}_{x,d\sim X,D}\left [d\log D_{I}(Q(x)) \right ] \right \}.
\end{equation}

To extract the domain-specific $Z_S$ defined in Definition~\ref{def:dom_spe}, we propose the use of the domain classifier $D_{\theta_{D_S}}$, that is trained to predict the domain label from $Z_S$. The corresponding parameters $\theta_{D_S}$ and $\theta_R$ are, therefore, optimized with the objective function below:
\begin{equation}
    \label{eq:zs_domain}
    \underset{\theta_{D_S},\theta_{R}}{\min} \left \{ L_{Z_{S}} := -\mathbb{E}_{x,d\sim X,D}\left [d\log D_{S}(R(x)) \right ]  \right \}.
\end{equation}

\textbf{Disentanglement between Domain-Invariant and Domain-Specific.}
The disentanglement condition between two random vectors $Z_I$ and $Z_S$ can be solved by forcing their covariance matrix, denoted by $\Cov(Z_I,Z_S)$ close to $0$. A detailed discussion of disentangled two representations is provided in Appendix~\ref{apd:discussion_disentangle}. The related parameters $(\theta_Q, \theta_R)$ are then updated in the following optimization problem:
\begin{equation}\label{eq:disentanglement}
    \underset{\theta_{Q}, \theta_{R}}{\min}\left \{ L_{D} := \mathbb{E}_{x\sim X}\left [ \left \| \Cov(Q(x), R(x)) \right \|_2 \right ]  \right \},
\end{equation}
where $\|\cdot\|_2$ is the $L_2$ norm.\\
\emph{Sufficiency of domain-specific and domain-invariant w.r.t. the classification task.} The goal of the classifier $F$ parameterized by $\theta_F$ is to predict the label of the original sample $X$ based on the domain-invariant $Z_I$ and domain-specific $Z_S$, i.e.,
\begin{equation}
    \label{eq:inference}
    \hat{Y} = F_{\theta_F}(Z_I \oplus Z_S),
\end{equation}
where $\oplus$ denotes the concatenation operation. Then, the training process of $F$ is then performed by solving
\begin{equation}
    \label{eq:class}
    \begin{array}{l}
    \underset{\theta_{Q},\theta_{R},\theta_{F}}{\min}\left \{ L_{T} := -\mathbb{E}_{x,y\sim X,Y}\left [y\log F(Q(x), R(x)) \right ] \right \}.
    \end{array}
\end{equation}

\textbf{Meta-Training for Domain-Specific Information.}
To encourage the domain-specific representation $Z_S$ to adapt information learned from the source domains to the unseen target domain, we introduce the use of meta-learning framework~\cite{li2017learning}, targeting a robust generalization. Note that the domain-invariant feature $Z_I$ remains during the meta-learning procedure. In particular, each source domain $S_m, \; m\in\overline{1,N}$ is split into two sub-domains, namely meta-train $S_{mr}$ and meta-test $S_{me}$. The domain-specific parameters $\theta_R$ and the classifier parameters $\theta_F$ are then jointly optimized as follows:
\begin{equation}
    \label{eq:meta-update}
    \begin{array}{l}
    \underset{w}{\min}\left \{ L_{T_{m}} := f\left ( w-\nabla f\left ( w,S_{mr} \right ), S_{me} \right ) \right \},
    \end{array}
\end{equation}
where $w = \left ( \theta_{R}, \theta_{F} \right )$ and
\begin{equation}
    \label{eq:meta-train}
    f\left ( w,S_m \right ) = -\mathbb{E}_{x,y\sim X,Y}\left [y\log F(Z_{I}, R(x)) \right ].
\end{equation}

\textbf{Training and Inference.} The pseudo-code for training and inference processes of our proposed \acrshort{DSDI} framework is presented in Algorithm~\ref{alg:algorithm}. Each iteration of the training process consists of two steps:
\begin{itemize}
    \item[i)] First, we integrate the objective functions~\eqref{eq:atloss}, \eqref{eq:zs_domain}, \eqref{eq:disentanglement} and \eqref{eq:class} to construct an  objective function $L_{A}$ defined as follows:
\begin{equation}
    \label{eq:e2e_loss}
    \begin{array}{l}
    \underset{\theta_{Q},\theta_{D_S},\theta_{R},\theta_{F}}{\min}\underset{\theta_{D_I}}{\max} \left \{ L_{A} := \lambda_{Z_{I}} L_{Z_{I}} + \lambda_{Z_S} L_{Z_S} + \lambda_{D} L_{D} +  L_{T} \right \},
    \end{array}
\end{equation}
where $\lambda_{Z_{I}}$, $\lambda_{Z_S}$ and $\lambda_{D}$ are selected as the balanced parameters.
    \item[ii)] The second step is to employ meta-training to adapt task-related domain-specific from source domains to unseen domains. In each mini-batch, the meta-train and meta-test are split, then the gradient transformation step from meta-train domains to the meta-test domain is performed by solving the optimization problem~\eqref{eq:meta-update}. 
\end{itemize}

\begin{algorithm}[ht!]
\caption{Training and Inference processes of \acrshort{DSDI}}
 \label{alg:algorithm}
\SetAlgoLined
\textbf{Training Input}: Source domain $S^{(i)}$, encoder $Q_{\theta_{Q}}$, $R_{\theta_{R}}$, domain classifier $D_{\theta_{D_I}}$, $D_{\theta_{D_S}}$ for $Z_I$, $Z_S$, task classifier $F_{\theta_{F}}$, batch size $B$, learning rate $\eta$. {\bfseries Output:} The optimal: $Q_{\theta_{Q}}^*$, $R_{\theta_{R}}^*$, $F_{\theta_{F}}^*$\;
\For{$\text{ite}=1\rightarrow \text{iterations}$}{
  Sample $S_{B}$ with a mini-batch $B$ for each domain $S^{(i)}$\;
  Compute $L_{A}$ using Eq.~\eqref{eq:e2e_loss} and perform gradient update $\nabla_{\theta_{Q}, \theta_{R}, \theta_{D_I}, \theta_{D_S}, \theta_{F}} L_{A}$ with $\eta$.\;
     \For{$\text{j}=1\rightarrow \text{N (number of source domains)}$}{
         Split Meta-train $S_{B/j}$, Meta-test $S_{j}$\;
         {\bfseries Meta-train}: Perform gradient update $\nabla_{\theta_{R}, \theta_{F}}$ by minimizing Eq.~\eqref{eq:meta-train} with $S_{B/j}$ and $\eta$\;
         {\bfseries Meta-test}: Compute $L_{T_{m}}$ using Eq.~\eqref{eq:meta-update} with $S_{j}$ and updated gradient from {\bfseries Meta-train}\;
         {\bfseries Meta-optimization}: Perform gradient update $\nabla_{\theta_{R}, \theta_{F}} L_{T_{m}}$ with $\eta$\;
     }
 }
 {\bfseries Inference Input:} Target domain $S^T$, optimal: $Q_{\theta_{Q}}^*$, $R_{\theta_{R}}^*$, $F_{\theta_{F}}^*$.
 {\bfseries Output:} $Y^T$ using Eq.~\eqref{eq:inference}\;
\end{algorithm}
\section{Experiments}
\subsection{Experimental settings}
\textbf{Dataset.}
To evaluate the effectiveness of the proposed method, we utilize 7 commonly used datasets including: \textbf{Colored-MNIST}~\cite{arjovsky2020irm}: includes 70000 samples of dimension $(2, 28, 28)$ in binary classification problem with noisy label, from MNIST over 3 domains with noisy rate $d \in $ \big\{0.1, 0.3, 0.9\big\}, \textbf{Rotated-MNIST}~\cite{ghifary2015object}: contains $70000$ samples of dimension $(1, 28, 28)$ and 10 classes, rotated from MNIST over 6 domains $d\in$ \big\{0, 15, 30, 45, 60, 75\big\}, \textbf{VLCS}~\cite{fang2013vlcs}: includes 10729 samples of dimension $(3, 224, 224)$ and 5 classes, over 4 photographic domains $d \in $ \big\{Caltech101, LabelMe, SUN09, VOC2007\big\}, \textbf{PACS}~\cite{li2017deeper}: contains 9991 images of dimension $(3, 224, 224)$ and 7 classes, over 4 domains $d \in $ \big\{artpaint, cartoon, sketches, photo\big\}, \textbf{Office-Home}~\cite{venkateswara2017deep}: has 15500 daily images of dimension $(3, 224, 224)$ and 65 categories, over 4 domains $d \in$ \big\{art, clipart, product, real\big\}, \textbf{Terra Incognita}~\cite{terra}: includes 24778 wild photographs of dimension $(3, 224, 224)$ and 10 animals, over 4 camera-trap domains $d \in$ \big\{L100, L38, L43, L46\big\}, and \textbf{DomainNet}~\cite{peng2019moment}: contains $586575$ images of dimension $(3, 224, 224)$ and 345 classes, over 6 domains  $d \in$ \big\{clipart, infograph, painting, quickdraw, real, sketch\big\}. The detail of each dataset is provided in Appendix~\ref{apd:dataset}.

\textbf{Baseline.}
Following \textbf{DomainBed}~\cite{gulrajani2020domainbed} settings, we compare our model with 14 related methods in DG which are divided by 5 common techniques, including: \textit{\textbf{Standard Empirical Risk Minimization}}: Empirical Risk Minimization (\textbf{ERM}~\cite{vapnik1998erm}); \textit{\textbf{domain-specific-learning}}: Group Distributionally Robust Optimization (\textbf{GroupDRO}~\cite{sagawa2020dro}), Marginal Transfer Learning (\textbf{MTL}~\cite{blanchard2011generalizing,blanchard2021mtl}), Adaptive Risk Minimization (\textbf{ARM}~\cite{zhang2021arm}); \textit{\textbf{Meta-learning}}: Meta-Learning for \acrshort{DG} (\textbf{MLDG}~\cite{li2017learning}); \textit{\textbf{domain-invariant-learning}}: Invariant Risk Minimization (\textbf{IRM}~\cite{arjovsky2020irm}), Deep CORrelation ALignment (\textbf{CORAL}~\cite{sun2016corall}), Maximum Mean Discrepancy (\textbf{MMD}~\cite{li2018mmd}), Domain Adversarial Neural Networks (\textbf{DANN}~\cite{ganin2016dann}), Class-conditional DANN (\textbf{CDANN}~\cite{li2018cdann}), Risk Extrapolation (\textbf{VREx}~\cite{krueger2021vrex}); \textit{\textbf{Augmenting data}}: Inter-domain Mixup (\textbf{Mixup}~\cite{xu2019adversarial,yan2020improve,wang2020heterogeneous}), Style-Agnostic Networks (\textbf{SagNets}~\cite{nam2021sagnets}), Representation Self Challenging (\textbf{RSC}~\cite{huang2020rsc}). The detail of each method is provided in Appendix~\ref{apd:baseline}.

We use the training-domain validation set technique as proposed in DomainBed~\cite{gulrajani2020domainbed} for model selection. In particular, for all datasets, we first merge the raw training and validation, then, we run the test three times with three different seeds. For each random seed, we randomly split training and validation and choose the model maximizing the accuracy on the validation set, then compute performance on the given test sets. The mean and standard deviation of classification accuracy from these three runs are reported. We evaluate generalization performance based on backbones MNIST-ConvNet~\cite{gulrajani2020domainbed} for MNIST datasets and ResNet-50~\cite{resnet} for non-MNIST datasets to compare with the mentioned methods. Data-processing techniques, model architectures, hyper-parameters, and changes of objective functions during training are presented in detail in Appendix~\ref{apd:implementation}~\ref{apd:loss}. All source code to reproduce results are available at \href{https://github.com/VinAIResearch/mDSDI}{\textit{https://github.com/VinAIResearch/mDSDI}}.

\subsection{Results}
\begin{table}[ht!]
\vskip -0.1in
\caption{Classification accuracy (\%) for all algorithms and datasets summarization. Our \acrshort{DSDI} method achieves highest accuracy on average when comparing 14 popular DG algorithms across 7 benchmark datasets.}
\label{tab:results}
\centering
\scalebox{0.77}{
\begin{tabular}{lcccccccc}
\toprule
\textbf{Method} & \textbf{CMNIST} & \textbf{RMNIST} & \textbf{VLCS} & \textbf{PACS} & \textbf{OfficeHome} & \textbf{TerraInc} & \textbf{DomainNet} & \textbf{Average} \\
\midrule
ERM~\cite{vapnik1998erm} & 51.5$\pm$0.1 & 98.0$\pm$0.0 & 77.5$\pm$0.4 & 85.5$\pm$0.2 & 66.5$\pm$0.3 & 46.1$\pm$1.8 & 40.9$\pm$0.1 & 66.6\\
IRM~\cite{arjovsky2020irm} & 52.0$\pm$0.1 & 97.7$\pm$0.1 & 78.5$\pm$0.5 & 83.5$\pm$0.8 & 64.3$\pm$2.2 & 47.6$\pm$0.8 & 33.9$\pm$2.8 & 65.4\\
GroupDRO~\cite{sagawa2020dro} & 52.1$\pm$0.0 & 98.0$\pm$0.0 & 76.7$\pm$0.6 & 84.4$\pm$0.8 & 66.0$\pm$0.7 & 43.2$\pm$1.1 & 33.3$\pm$0.2 & 64.8\\
Mixup~\cite{xu2019adversarial,yan2020improve,wang2020heterogeneous} & 52.1$\pm$0.2 & 98.0$\pm$0.1 & 77.4$\pm$0.6 & 84.6$\pm$0.6 & 68.1$\pm$0.3 & 47.9$\pm$0.8 & 39.2$\pm$0.1 & 66.7\\
MLDG~\cite{li2017learning} & 51.5$\pm$0.1 & 97.9$\pm$0.0 & 77.2$\pm$0.4 & 84.9$\pm$1.0 & 66.8$\pm$0.6 & 47.7$\pm$0.9 & 41.2$\pm$0.1 & 66.7\\
CORAL~\cite{sun2016corall} & 51.5$\pm$0.1 & 98.0$\pm$0.1 & 78.8$\pm$0.6 & 86.2$\pm$0.3 & 68.7$\pm$0.3 & 47.6$\pm$1.0 & 41.5$\pm$0.1 & 67.5\\
MMD~\cite{li2018mmd} & 51.5$\pm$0.2 & 97.9$\pm$0.0 & 77.5$\pm$0.9 & 84.6$\pm$0.5 & 66.3$\pm$0.1 & 42.2$\pm$1.6 & 23.4$\pm$9.5 & 63.3\\
DANN~\cite{ganin2016dann} & 51.5$\pm$0.3 & 97.8$\pm$0.1 & 78.6$\pm$0.4 & 83.6$\pm$0.4 & 65.9$\pm$0.6 & 46.7$\pm$0.5 & 38.3$\pm$0.1 & 66.1\\
CDANN~\cite{li2018cdann} & 51.7$\pm$0.1 & 97.9$\pm$0.1 & 77.5$\pm$0.1 & 82.6$\pm$0.9 & 65.8$\pm$1.3 & 45.8$\pm$1.6 & 38.3$\pm$0.3 & 65.6\\
MTL~\cite{blanchard2011generalizing,blanchard2021mtl} & 51.4$\pm$0.1 & 97.9$\pm$0.0 & 77.2$\pm$0.4 & 84.6$\pm$0.5 & 66.4$\pm$0.5 & 45.6$\pm$1.2 & 40.6$\pm$0.1 & 66.2\\
SagNets~\cite{nam2021sagnets} & 51.7$\pm$0.0 & 98.0$\pm$0.0 & 77.8$\pm$0.5 & \textbf{86.3$\pm$0.2} & 68.1$\pm$0.1 & \textbf{48.6$\pm$1.0} & 40.3$\pm$0.1 & 67.2\\
ARM~\cite{zhang2021arm} & \textbf{56.2$\pm$0.2} & \textbf{98.2$\pm$0.1} & 77.6$\pm$0.3 & 85.1$\pm$0.4 & 64.8$\pm$0.3 & 45.5$\pm$0.3 & 35.5$\pm$0.2 & 66.1\\
VREx~\cite{krueger2021vrex} & 51.8$\pm$0.1 & 97.9$\pm$0.1 & 78.3$\pm$0.2 & 84.9$\pm$0.6 & 66.4$\pm$0.6 & 46.4$\pm$0.6 & 33.6$\pm$2.9 & 65.6\\
RSC~\cite{huang2020rsc} & 51.7$\pm$0.2 & 97.6$\pm$0.1 & 77.1$\pm$0.5 & 85.2$\pm$0.9 & 65.5$\pm$0.9 & 46.6$\pm$1.0 & 38.9$\pm$0.5 & 66.1\\
\acrshort{DSDI} (Ours)& 52.2$\pm$0.2 & 98.0$\pm$0.1 & \textbf{79.0$\pm$0.3} & 86.2$\pm$0.2 & \textbf{69.2$\pm$0.4} & 48.1$\pm$1.4 & \textbf{42.8$\pm$0.1} & \textbf{67.9}\\
\bottomrule
\end{tabular}}
\end{table}

Table~\ref{tab:results} summarizes the results of our experiments on 7 benchmark datasets when compared with mentioned methods. The full result per dataset and domain is provided in Appendix~\ref{apd:results}. From these results, we draw three conclusions about our \acrshort{DSDI} model:

\textbf{Our \acrshort{DSDI} still preserves domain-invariant information.}
We observe in some target domains which have background-less images and assume only contain domain-invariant information such as Colored-MNIST, Rotated-MNIST, or Terra Incognita (similar observation in cartoon or sketch in PACS, clip-art or product in OfficeHome, and quickdraw in DomainNet. Full results in Appendix~\ref{apd:results}), our \acrshort {DSDI} model still achieves competitive results with other baselines (e.g., $52.2\%$ in Colored-MNIST, $98.0\%$ in Rotated-MNIST, and $48.1$\% in Terra Incognita) which are based on domain-invariant-learning techniques such as DANN, C-DANN, CORAL, MMD, IRM, and VREx. Those results demonstrate the effectiveness of our adversarial training technique for extracting domain-invariant features. Furthermore, due to considering disentangled domain-invariant and domain-specific latent, even in situations where the samples do not have domain-specific, our model still performs well by retaining informative domain-invariance.

\textbf{Our \acrshort{DSDI} could capture the usefulness of domain-specific information.}
In contrast, we observe that in some target domains that have relevant domain-specific with source domains such as landscape background of the object class from photographic pictures in VLCS (similar observation in PACS such as dogs in the yard or guitars lying on a table in photo and art domain, bed in the room or bike parked on the street in the Art and Real-world domain of OfficeHome. Full results in Appendix~\ref{apd:results}), \acrshort{DSDI} achieves significantly higher results than other methods (e.g., $79.0\%$ in VLCS, $86.2\%$ in PACS). This means that our domain-invariant features not only support generalization better but also our domain-specific ones cover helpful information in special scenarios such as backgrounds and colors related to objects in the classification task. Moreover, when comparing with other domain-specific based techniques such as GroupDRO, MTL, and ARM, the results showed that domain-specific features learned by meta-training from our model are more helpful than theirs and have captured useful domain-specific features from those object-background relations.

\textbf{Extending beyond the invariance view to usefulness domain-specific information is important.}
As shown in Table~\ref{tab:results}, our \acrshort{DSDI} has the highest average number with $67.9\%$ (highlighted with statistically significant according to a $t$-test at a significance level $\alpha = 0.05$). The reason why our model outperforms other baselines could be explained by the fact that their domain-invariant methods are not able to capture domain-specific information, and so have poor performance. Meanwhile, when comparing with other domain-specific based methods, their models only concentrate on domain-specific techniques, and so provide inferior domain-invariant information to our techniques in some background-less images. In contrast, due to considering disentangled domain-invariant and domain-specific features, and having the right strategy to learn each latent, our model captures both this useful information, hence, outperforms their results. Not only has the highest average number, but our method also dominates other methods on a known large-scale dataset such as $69.2\%$ in Office-Home or $42.8\%$ in DomainNet. This implies that besides the essential combination between domain-invariant and domain-specific, when the number of datasets increases, our method can extract more relevant information for complex tasks, such as classifying 345 classes in DomainNet. These results also mean that our model has a balance between informative domain-invariant and domain-specific features to adapt better to different environments than others, therefore showing the highest average in all settings.

\subsection{How does \acrshort{DSDI} work?} 
\begin{figure}[ht!]
\begin{center}
   \includegraphics[width=1.0\linewidth]{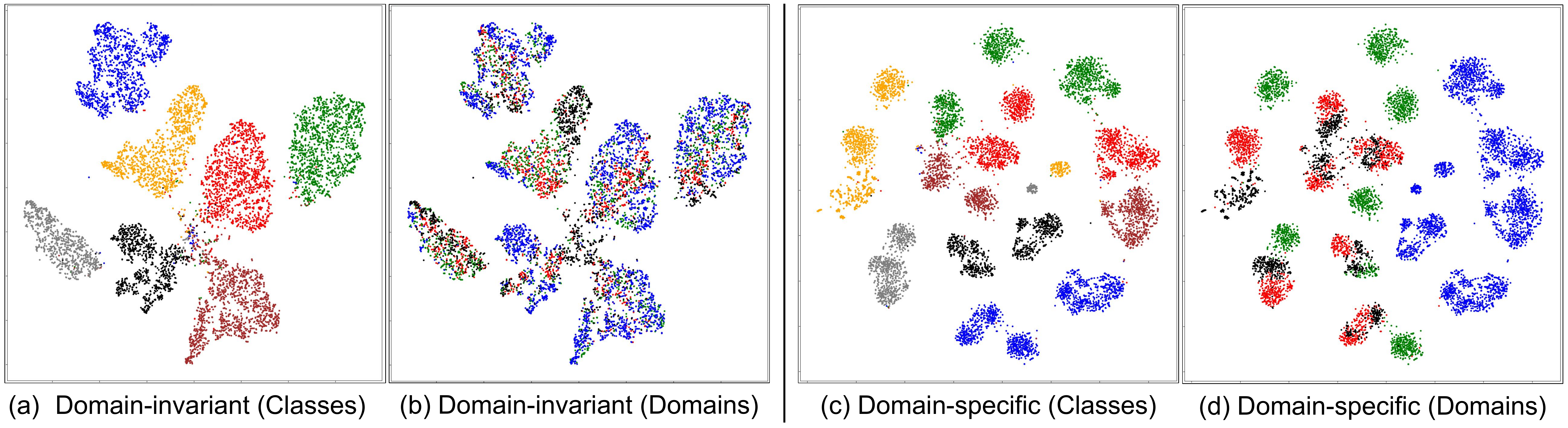}
   \caption{Feature visualization for domain-invariant: (a): different colors represent different classes; (b): different colors indicate different domains. Feature visualization for domain-specific: (c): different colors represent different classes; (d): different colors indicate different domains. Source domain includes: art (red), cartoon (green), sketch (blue) while target domain is photo (black) in the domain plots. Best viewed in color (Zoom in for details).}
\label{fig:tSNE-fig}
\end{center}
\vskip -0.2in
\end{figure}

To better understand our framework, we visualize the distribution of the learned features with t-SNE~\cite{vandermaaten08a} to analyze the feature space of domain-invariant and domain-specific. As shown in Figure~\ref{fig:tSNE-fig} on PACS Dataset, our domain-invariant  extractor can minimize the distance between the distribution of the domains (see Figure~\ref{fig:tSNE-fig}: (b)). However, these domain-invariant features still make mistakes on the classification task, indicated by a mixture of points from different class labels in the middle (see Figure~\ref{fig:tSNE-fig}: (a)), and many of these points are from the target domain (black color in the Figure~\ref{fig:tSNE-fig}: (b)). Meanwhile, the domain-specific  representation better distinguishes points by class label (see Figure~\ref{fig:tSNE-fig}: (c)). More importantly, the photo domain's specific features (black) are close to the art domain (red) (see Figure~\ref{fig:tSNE-fig}: (d)). This is reasonable because only these two domains include backgrounds related to the object class. It implies that meta-training in our model well learns specific features that can be adapted to the new unseen domain.

\subsection{Ablation study: Important of \acrshort{DSDI} on the Background-Colored-MNIST dataset}\label{ablation}
This section examines our system design by checking its performance under different settings in the real scenario with our generated dataset (a similar experiment with PACS benchmark dataset is in Appendix~\ref{apd:ablation}).

\textbf{Background-Colored-MNIST dataset.}
Figure~\ref{fig:colored-mnist-fig} illustrates our Background-Colored-MNIST, generated from the original MNIST. We assume the domain-invariant is the digit's sketch and design the dataset so that the background color is domain-specific. As a result, on the unseen domain, domain-specific will be useful for the classification task. Specifically, the dataset includes three source domains $d_{tr}$, different by digit's color \big\{red, green, blue\big\}, generated from a subset with 1000 training images for MNIST per each domain. In each source domain, the background color is the same for intra-class images but different across classes. In the target domain, 10000 testing images of MNIST are colored for one target domain $d_{te}$ with digit color \big\{orange\big\}. In this domain, each class's background color is similar to the same class's background color in one of three source domains.

\textbf{Importance of \acrshort{DSDI}.}
We aim to prove the combination of learning disentangled representation domain-specific, domain-invariant, and meta-training on domain-specific are important in this scenario. To do so, we compare our model under nine settings: learning domain-invariant only (DI), learning domain-specific only (DS), meta-training on domain-invariant (DI-Meta), meta-training on domain-specific (DS-Meta), a combination of domain-invariant and domain-specific without disentanglement loss $L_D$ (DSDI-Without $L_D$), a combination of domain-invariant and domain-specific without meta-training (DSDI-Without Meta), meta-training on both representation $Z_I$ and $Z_S$ (DSDI-Meta), meta-training on domain-invariant without domain-specific (DSDI-Meta DI) and our proposed framework (\acrshort{DSDI}-Meta DS), which is meta-training on domain-specific without domain-invariant.\\
Table~\ref{tab:colored-mnist} shows that our model is the best setting with $89.7\%$. It proves that combining domain-invariant and domain-specific is crucial by dominating the settings with only domain-invariant or domain-specific (DI, DI-Meta, DS, DS-Meta). Regarding disentangling two representations $Z_I$ and $Z_S$, it is worth noting that without disentanglement loss $L_D$, the model only achieves $81.4\%$, which is lower than \acrshort{DSDI}. It reveals adding disentanglement loss $L_D$ is essential to boost our model performance. Compared with meta-training on both, which only reaches $82.1\%$ and on domain-invariant are $79.0\%$, it implies that meta-training is necessary but only for the domain-specific, and our arguments are reasonable. It also shows that our combination with domain-invariant and domain-specific is not easy: a deep ensemble between two neural networks, but existing a deep-down insight in our framework in \acrshort{DG}, when compared with \acrshort{DSDI}-Without Meta which can be seen as a type of ensemble, is only around $80.4\%$.

\begin{minipage}{\textwidth}
  \begin{minipage}[hb]{0.49\textwidth}
    \centering
    \includegraphics[width=1.0\linewidth]{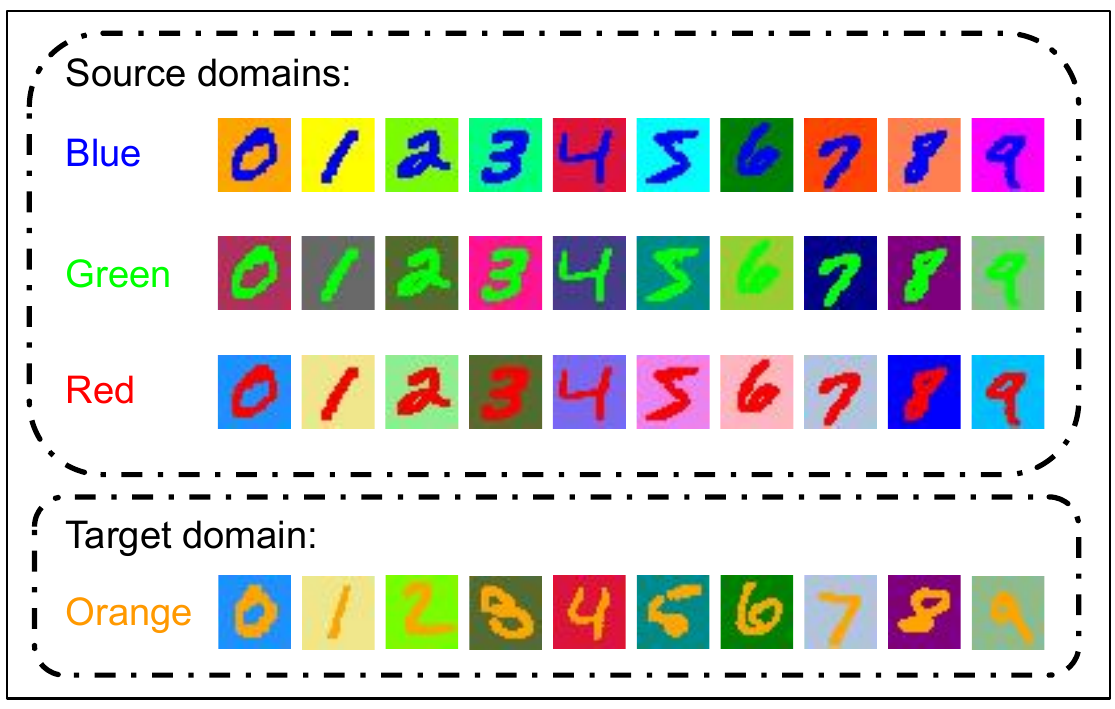}
    \captionof{figure}{Background-Colored-MNIST Dataset, where source domains include \big\{red, green, blue\big\} digit colors and target domain has \big\{orange\big\} color.}
    \label{fig:colored-mnist-fig}
  \end{minipage}
  \hfill
  \begin{minipage}[hb]{0.5\textwidth}
    \centering
    \captionof{table}{Classification accuracy (\%) on Background-Colored-MNIST. Ablation study shows impact of domain-invariant when combined with meta-training on domain-specific in our method.}
    \label{tab:colored-mnist}
    \begin{tabular}{lc}
        \toprule
        \textbf{Method} & \textbf{Accuracy}\\
        \midrule
        DI & 65.7$\pm$4.6\\
        DI-Meta & 63.6$\pm$5.1\\
        DS & 70.7$\pm$4.8\\
        DS-Meta & 75.3$\pm$3.4\\
        DSDI-Without $L_D$ & 81.4$\pm$2.6\\
        DSDI-Without Meta & 80.4$\pm$1.7\\
        DSDI-Meta & 82.1$\pm$1.4\\
        DSDI-Meta DI & 79.0$\pm$2.3\\
        \acrshort{DSDI}-Meta DS (Ours) & \textbf{89.7$\pm$0.8}\\
        \bottomrule
    \end{tabular}
    \end{minipage}
\end{minipage}
\vskip -0.1in
\section{Conclusion and Discussion}
Despite being aware of the importance of domain-specific information, little investigation into the theory and a rigorous algorithm to explore its representation. To the best of our knowledge, our work provides the first theoretical analysis to understand and realize the efficiency of domain-specific information in domain generalization. The domain-specific contains unique characteristics and when combined with domain-invariant information can significantly aid performance on unseen domains. Following our theoretical insights based on the information bottleneck principle, we propose a \acrshort{DSDI} algorithm which disentangles these features. We next introduce the use of the meta-training scheme to support domain-specific to adapt information from source domains to unseen domains. Our experimental results demonstrate \acrshort{DSDI} brings out competitive results with related approaches in domain generalization. In addition, the ablation study with our Background-Colored-MNIST further illustrated and demonstrated the efficiency of combining domain-invariant and domain-specific via our proposed \acrshort{DSDI}. Our theoretical analysis and proposed \acrshort{DSDI} framework can facilitate fundamental progress in understanding the behavior of both domain-invariant and domain-specific representation in domain generalization.

Toward a robustness algorithm that can effectively learn both domain-invariant and domain-specific features, there are certainly many challenges that remain in our paper, for example, a theorem to explain when domain-specific may hurt performance in the unseen domain, a stronger connection between theory in implementation, a method to make two representations to be non-linearly independent as well as a lower computational cost of the covariance matrix. In the future, we plan to continue tackling these challenges to provide a better understanding and learning framework in domain generalization then extending to broader settings of transfer learning. A detailed clarification, discussion, and plausible methods in the future work are provided in Appendix~\ref{apd:discussion}.

\bibliographystyle{unsrt}
\bibliography{neurips_2021}

\newpage
\appendix
\section*{\centering\textbf{Exploiting Domain-Specific Features to Enhance Domain Generalization (Supplementary Material)}}

In this supplementary material, we collect proofs and remaining materials that were deferred from the main paper. In Appendix~\ref{apd:proof}, we provide the proofs for all the results in the main paper, including: proof of Lemma~\ref{lem:determinism} in Appendix~\ref{proof:lemma}, proof of Theorem~\ref{theo:inv} in Appendix~\ref{proof:theorem}. In Appendix~\ref{apd:discussion}, we discuss when domain-specific will effect negatively on the unseen target domain in~\ref{apd:discussion_ds}, clarify the gap between the presented theory and the implemented algorithm in~\ref{apd:discussion_gap}, and continue to discuss the disentanglement technique as well as potential future directions in~\ref{apd:discussion_disentangle}. In Appendix~\ref{apd:exp}, we provide additional information about our experiments, including: sufficient details about the dataset in Appendix~\ref{apd:dataset}, baseline details in Appendix~\ref{apd:baseline}, implementation details to reproduce our experiments in Appendix~\ref{apd:implementation}, further analysis about the results for each benchmark dataset in Appendix~\ref{apd:results}, our model’s behaviour when training in Appendix~\ref{apd:loss}, and additional ablation study on the benchmark dataset in Appendix~\ref{apd:ablation}.

\section{Proofs}\label{apd:proof}
\subsection{Proof of Lemma~\ref{lem:determinism}}\label{proof:lemma}
\begin{proof}
When $Z_{X^1}$ is output of deterministic functions from $X^1$, for any A in the sigma-algebra induced by $Z_{X^1}$ we have 
$$\mathbb{E}[\mathbbm{1}_{Z_{X^1}\in A}|X^1,\left \{ Y, X^2 \right \}]=\mathbb{E}[\mathbbm{1}_{Z_{X^1}\in A}|X^1,X^2]=\mathbb{E}[\mathbbm{1}_{Z_{X^1}\in A}|X^1],$$
which implies $Y\perp \!\!\! \perp Z_{X^1}|X^1$ and $X^2 \perp \!\!\! \perp Z_{X^1}|X^1$.
\end{proof}

\subsection{Proof of Theorem~\ref{theo:inv}}\label{proof:theorem}
\begin{proof}
The proofs contain two parts. The first one is showing the results for the label-related learned representations and the second one is for the domain-invariant learned representations.

\textit{\underline{Part (1). Label-related Learned Representations:}} Adapting the Data Processing Inequality (DPI by~\cite{cover2006}) in the Markov chain $X^2 \leftrightarrow Y \leftrightarrow X^1 \rightarrow Z_{X^1}$ (Lemma~\ref{lem:determinism}), $I(Z_{X^1};Y)$ is maximized at $I(X^1;Y)$.\\ 
Since both label-related learned presentation ($Z_{sup}$ and $Z_{sup^*}$) maximize $I(Z_{X^1};Y)$, we conclude $$I(Z_{sup};Y)=I(Z_{sup^*};Y)=I(X^1;Y).$$

\textit{\underline{Part (2). Domain-invariant Learned Representations:}} First, we have
$$I(Z_{X^1};X^2)=I(Z_{X^1};Y)-I(Z_{X^1};Y|X^2)+I(Z_{X^1};X^2|Y)=I(Z_{X^1};Y;X^2)+I(Z_{X^1};X^2|Y)$$ and $$I(X^1;X^2)=I(X^1;Y)-I(X^1;Y|X^2)+I(X^1;X^2|Y)=I(X^1;Y;X^2)+I(X^1;X^2|Y).$$\\
By DPI in the Markov chain $X^2 \leftrightarrow Y \leftrightarrow X^1 \rightarrow Z_{X^1}$ (Lemma~\ref{lem:determinism}), we know
\begin{itemize}
    \item $I(Z_{X^1};X^2)$ is maximized at $I(X^1;X^2)$
    \item $I(Z_{X^1};X^2;Y)$ is maximized at $I(X^1;X^2;Y)$ 
    \item $I(Z_{X^1};X^2|Y)$ is maximized at $I(X^1;X^2|Y)$
\end{itemize}
Since both domain-invariant learned representation ($Z_{I}$ and $Z_{I^*}$) maximize $I(Z_{X^1};X^2)$, we have $$I(Z_{I};X^2)=I(Z_{I^*};X^2)=I(X^1;X^2).$$
Hence, $$I(Z_{I};X^2;Y)=I(Z_{I^*};X^2;Y)=I(X^1;X^2;Y)$$ and $$I(Z_{I};X^2|Y)=I(Z_{I^*};X^2|Y)=I(X^1;X^2|Y).$$

Using the result $I(Z_{I};X^2;Y)=I(Z_{I^*};X^2;Y)=I(X^1;X^2;Y)$, we get
$$I(Z_{I};Y)=I(X^1;Y)-I(X^1;Y|X^2)+I(Z_{I};Y|X^2)$$ and $$I(Z_{I^*};Y)=I(X^1;Y)-I(X^1;Y|X^2)+I(Z_{I^*};Y|X^2).$$ 

Since $I(Z_{X^1};X^1)=I(Z_{X^1};X^1|X^2)+I(Z_{X^1};X^1;X^2)$, where $I(Z_{X^1};X^1;X^2)=I(X^1;X^2)$ (Lemma~\ref{lem:determinism}) and $I(Z_{X^1};X^2)$ is maximized at $Z_{I^*}$. Then, $$I(Z_{X^1};X^1|X^2)=H(Z_{X^1}|X^2)-H(Z_{X^1}|X^1,X^2),$$
where $H(Z_{X^1}|X^2)$ contains no randomness (no information) as $Z_{X^1}$ being deterministic from $X^1$. Hence, minimizing $I(Z_{X^1};X^1|X^2)$ equivalents to minimizing $H(Z_{X^1}|X^2)$.

Using the result $I(Z_{I^*};Y)=I(X^1;Y)-I(X^1;Y|X^2)+I(Z_{I^*};Y|X^2)$ and due to $H(Z_{X^1}|X^2)$ is minimized at $Z_{I^*}$, $I(Z_{I^*};Y|X^2)=0$, we get $$I(Z_{I^*};Y)=I(X^1;Y)-I(X^1;Y|X^2)$$ or $$I(X^1;Y)=I(Z_{I^*};Y)+I(X^1;Y|X^2).$$

Combining the result in \textit{Part (1).} $I(Z_{sup};Y)=I(Z_{sup^*};Y)=I(X^1;Y)$ and $\epsilon_1=I(X^1;Y|X^2)$ (Assumption~\ref{asm:specific}), then
$$I(Z_{sup^*};Y)=I(Z_{I^*};Y)+\epsilon_1.$$

Moreover, by assuming that $\epsilon_1>0$ (Assumption~\ref{asm:specific}), we obtain
$$I(X^1;Y)=I(Z_{sup};Y)=I(Z_{sup^*};Y)=I(Z_{I^*};Y)+\epsilon_1>I(Z_{I^*};Y).$$
\end{proof}

\section{Further discussion}\label{apd:discussion}
\subsection{Domain-specific features may hurt classification performance in the test domain}\label{apd:discussion_ds}
Although our theorem~\ref{theo:inv} suggests that the representation learned by incorporating both domain-invariant and domain-specific has a stronger dependence on labels than that learned by only considering domain-invariant information on source domains $X^1$ and $X^2$. However, there is no guarantee that such representation can generalize well to the "unseen" target domain $X^T$. In the setting of domain generalization, as no target domain data is accessible, the domain-invariant feature may be "invariant" to the source domains only. Moreover, the domain-specific features of the target domain are not able to be extracted accurately if they correlate with other classes than they did in source domains.

In particular, the domain-specific features may hurt performance if they correlate with a different class label in the unseen domain than they did in the source domains (i.e., $I(Z_{I^*},Y)+\epsilon_1\neq I(X^T;Y^T)$). Following the example about cow and camel in~\cite{arjovsky2020irm}, if in an unseen domain, a camel stands in a field rather than desert, it will hurt our model performance. Another observation is in our ablation study that if the yellow background of number “2” in the source domain becomes the background of number “1” in the target domain, our model’s classification accuracy on the class number “1” will drop by 40\% in the target domain.

However, the domain-specific features likely will help classification performance in the real world (i.e., $I(Z_{I^*},Y)+\epsilon_1 = I(X^T;Y^T)$). The correlation between different class labels in the unseen than they did in the source domain is unlikely to appear. For example, camels are more likely to appear in the desert than in a field, and cows are more likely to appear in the field than in the desert. Moreover, our empirical results show several observations in the real world dataset in which domain-specific features of source domains strongly correlate with a class label in both source and target domains. In such case, our \acrshort{DSDI} always have positive results when compared to other domain-invariant based methods, showing our framework is able to select relevant domain-specific information from source domains to generalize well on the target domain.

\subsection{From theory to a unified framework}\label{apd:discussion_gap}
Despite not exactly minimizing measures of mutual information as discussed in information theory (which causes a high computational cost in \acrshort{DG}), our implementation still has a strong connection with the theory section, following these reasons:

\begin{itemize}
\item The classifier within standard cross-entropy minimization in Eq.~\eqref{eq:zs_domain} and Eq.~\eqref{eq:class} has been shown to be similar to minimal and sufficient representation with the label (definition~\ref{def:min_suf}) in~\cite{tsai2021selfsupervised}. This is also a fundamental definition of the information bottleneck method, which is proved in~\cite{tishby99information}.
\item The domain classifiers within an adversarial training framework in Eq.~\eqref{eq:atloss} can be used as a proxy for the minimal and sufficient representations with domain-invariant in definition~\ref{def:min_inv}. Because following the definition~\ref{def:min_inv}, it maximizes mutual information across source domains and minimizes redundancy specific information in a particular domain. Its optimal solution is similar to the adversarial training framework in Eq.~\eqref{eq:atloss}.
\item In the disentanglement loss in Eq.~\eqref{eq:disentanglement}, minimizing Covariance between two random variables is equivalent to minimizing mutual information between them. The reason is: we can derive $$\min I(Z_I,Z_S)=\min D_{KL}\left ( P(Z_I,Z_S),P(Z_I)P(Z_S) \right ).$$ Two random variables X, Y are disentangled (independent) if covariance (X, Y) = 0. Hence, if they are feature vectors $Z_I=\left [ Z_{I^1}, Z_{I^2},...Z_{I^m} \right ]$ and $Z_S=\left [ Z_{S^1}, Z_{S^2},...Z_{S^n} \right ]$, they will be disentangled (independence) when $\Cov(Z_{I^i},Z_{S^j})=0$ for every $(i, j)$-component in the covariance matrix.
\end{itemize}

\subsection{Disentangled two representations}\label{apd:discussion_disentangle}
The higher the feature dimensionality, the more expensive it is for calculating the Covariance matrix. Specifically, in an experiment in the PACS dataset, if the backbone is Resnet50, we have to store a $2048\times2048$ covariance matrix which causes our model to consume around 30GB (GPU).

Given this weakness, we tried an alternative solution with adversarial training to minimize the mutual information between $Z_I$ and $Z_S$. The idea is that we could derive $\min I(Z_I,Z_S)=\min D_{KL}(P(Z_I,Z_S),P(Z_I)P(Z_S))$, and if we shuffled w.r.t. to the index of samples in each mini-batch, we also could disentangle these features without being affected by feature dimension. More specifically, if we create a discriminator $D_{IS}$ to classify representation from $P(Z_I,Z_S)$ as fake samples and representations from $P(Z_I)P(Z_S)$ as real samples. Samples from $P(Z_I,Z_S)$ are obtained by passing the sample $X$ through the encoders $Q$ and $R$ to extract $(Z_I,Z_S)$. Samples from $P(Z_I)P(Z_S)$ are obtained by shuffling the exclusive representations of a batch of samples from $P(Z_I,Z_S)$. The encoder function $Q$ strives to generate exclusive representations $Z_I$ that combined with $Z_S$ look like drawn from $P(Z_I)P(Z_S)$. By minimizing the following objective function:
$$L_{D^{adv}}=\mathbb{E}_{P(Z_I)P(Z_S)}\left [\log D_{IS}(Z_I,Z_S)  \right ] + \mathbb{E}_{P(Z_I,Z_S)}\left [\log (1-D_{IS}(Z_I,Z_S)) \right ].$$
This is equivalent to minimizing the Jensen-Shannon divergence $D_{JS}(P_{Z_I,Z_S}||P(Z_I)P(Z_S))$ and thus the mutual information between $Z_I$ and $Z_S$ is minimized. However, since this is adversarial training, the results had a higher variance in terms of accuracy, so we will try to investigate more in future work.
\section{Additional experiments}\label{apd:exp}
\subsection{Dataset details}\label{apd:dataset}
\begin{figure}[ht!]
\begin{center}
  \includegraphics[width=1.0\linewidth]{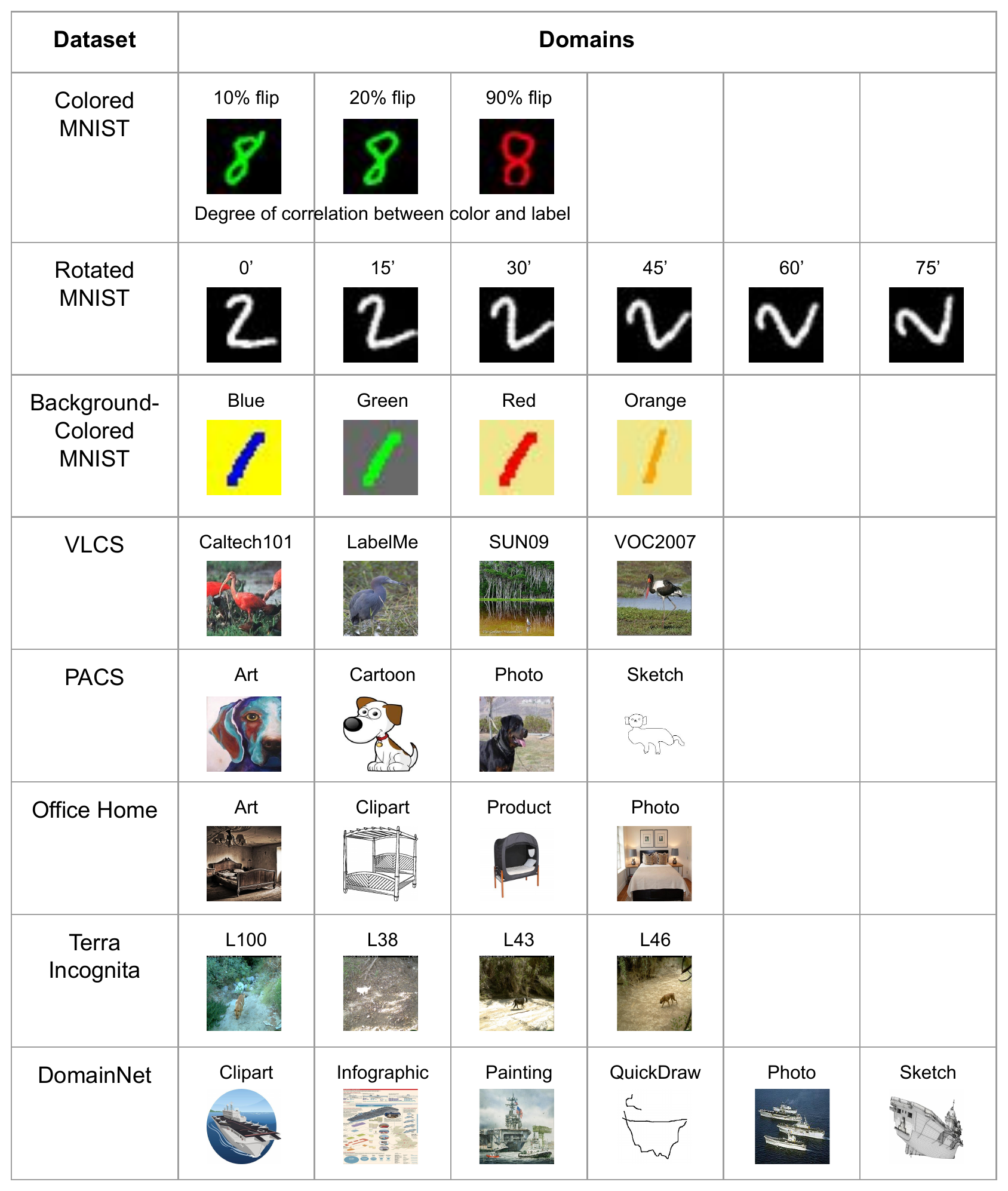}
\end{center}
  \caption{The benchmark dataset summarizations. For each dataset, we pick a single class and show illustrative images from each domain.}
\label{fig:dataset}
\end{figure}

This appendix provides more detail about the dataset in the main paper. Figure~\ref{fig:dataset} visualizes examples for each domain per dataset used in our experiments, including a totally of 7 image datasets widely used for classification tasks in \acrshort{DG} and our generated dataset Background-Colored-MNIST in Ablation Study~\ref{ablation}:
\begin{itemize}
\item{\textbf{Colored-MNIST~\cite{arjovsky2020irm}}} includes $70000$ samples of dimension $(2, 28, 28)$ in binary classification problem with noisy label, from MNIST over 3 domains with noisy rate $d\in$ \big\{0.1, 0.2, 0.9\big\}. The noisy rate is the correlation ratio between digit and color label. In particular, the construction including assigning a preliminary binary label $\hat{y}$ to the image based on the digit: $\hat{y} = 0$ for digits 0-4 and $\hat{y} = 1$ for 5-9, flipping $\hat{y}$ with probability 0.25 to obtain the final label $y$, flipping $y$ with probability $d$ corresponding to the noisy rate to obtain the color id $z$, and coloring the image red if $z = 1$ or green if $z = 0$. By doing so, if the noise rate in 2 source domains is $d= \left \{ 0.1,0.2 \right \}$ and $d=0.9$ in the test domain (correlation is reversed in the test environment), then this dataset will allow measuring the invariant learning ability of the model in which the actual invariant is the digit, and the color is just noisy information for the fooling model in source domains.

\item{\textbf{Rotated-MNIST~\cite{ghifary2015object}}} contains $70000$ samples of dimension $(1, 28, 28)$ with 10 classes per each domain, rotated from MNIST over 6 domains $d\in$ \big\{0, 15, 30, 45, 60, 75\big\}. This data set does not contain much domain-specific information since the background of binary images is black while only the sketch of the digit is rotated. And so, we assume the domain-invariant is the sketch of the digit.

\item{\textbf{Background-Colored-MNIST (Ours)}} contains 1000 training samples of dimension $(3, 28, 28)$ and 10 classes from MNIST per each source domain, colored by digit’s color over 3 domains $d_{tr} \in$ \big\{red, green, blue\big\}. In addition, the background color is the same for intra-class images but different across classes. There are 10000 testing samples in the target domain, which are colored by $d_{te} \in$ \big\{orange\big\} and each class’s background color is similar to the same class’s background color in one of three source domains. This data set is used for our ablation study, with the assumption that domain-invariant is the digit’s sketch and domain-specific is the background color.

\item{\textbf{VLCS~\cite{fang2013vlcs}}} includes 10729 samples of dimension $(3, 224, 224)$ and 5 classes, over 4 photographic domains $d \in $ \big\{Caltech101, LabelMe, SUN09, VOC2007\big\}. The samples from this dataset are collected by taking pictures from real life, hence, contain a lot of domain-specific information on the background such as the landscape of streets where the cars are parked or fields where the bird is eating. However, we observe that in this dataset, besides samples that only contain the object of the class (assume domain-invariant information) without informative backgrounds such as zoom of car or bird, some samples only contain backgrounds (assume domain-specific information) such as houses, ocean or sky landscape.

\item{\textbf{PACS~\cite{li2017deeper}}} contains 9991 images of dimension $(3, 224, 224)$ and 7 classes, over 4 domains $d \in $ \big\{artpaint, cartoon, sketches, photo\big\}. This is one of the most popular benchmark data sets in \acrshort{DG}, while in the artpaint and photo domain contains colored backgrounds that are specific and correlate with the label such as dogs in the yard, the cartoon and sketches only have a white background. And so, we assume the domain-invariant is the object of class and domain-specific is the color and background.

\item{\textbf{Office-Home~\cite{venkateswara2017deep}}} has 15588 daily images of dimension $(3, 224, 224)$ and 65 categories, over 4 domains $d \in$ \big\{art, clipart, product, real\big\}. Similar to PACS, the majority of images in two domains clipart and product do not include much informative domain-specific information such as color, backgrounds which is related to objects and we assume the domain-invariant is the sketch of the object. Meanwhile, the art and real domains contain more domain-specific features, for instance, the bed usually in a room.

\item{\textbf{Terra Incognita}~\cite{terra}} includes $24788$ wild photographs of dimension $(3, 224, 224)$ with 10 animals, over 4 camera-trap domains $d\in$ \big\{L100, L38, L43, L46\big\}. This dataset contains photographs of wild animals taken by camera traps; camera trap locations are different across domains. Since these cameras are static, different animals still have the same background, or in other words, there is no correlation between background and animal label in this dataset. Hence, lacks domain-specific features and mainly contains domain-invariant information of the object animal.

\item{\textbf{DomainNet~\cite{peng2019moment}}} contains $596006$ images of dimension $(3, 224, 224)$ and 345 classes, over 6 domains $d \in$ \big\{clipart, infograph, painting, quickdraw, real, sketch\big\}. This is the biggest dataset in terms of the number of samples and classes. The two domains: quickdraw and sketch, only contain conceptual drawings of the object, and so, we assume only having domain-invariant. In contrast, 4 domains: clipart, infographic, painting, and photo have more domain-specific information such as colors, backgrounds, text to describe the object.
\end{itemize}

\subsection{Baseline details}\label{apd:baseline}
This appendix provides an exhaustive literature review about 14 related methods in \acrshort{DG} which are used to make comparisons with our model, divided by 5 common techniques:
 
\textit{\textbf{Standard Empirical Risk Minimization}}: 
\begin{itemize}
\item{Empirical Risk Minimization (\textbf{ERM}~\cite{vapnik1998erm})} aggregates all the source domain data together, minimized with cross-entropy for classification loss.
\end{itemize}

\textit{\textbf{Domain-specific learning}}: 
\begin{itemize}
\item{Group Distributionally Robust Optimization (\textbf{GroupDRO}~\cite{sagawa2020dro})} performs ERM while increasing the importance of domains by weighing mini-batches of the training distribution proportional with larger errors.

\item{Marginal Transfer Learning (\textbf{MTL}~\cite{blanchard2011generalizing,blanchard2021mtl})} estimates a kernel mean embedding per domain, passed as a second argument to the classifier. Then, these embeddings are estimated using single test examples at test time (only applicable when using RKHS-based learners).

\item{Adaptive Risk Minimization (\textbf{ARM}~\cite{zhang2021arm})} an extension of MTL where a separate CNN computes the domain embedding, appended to the input images as additional channels.

\end{itemize}

\textit{\textbf{Meta-learning}}: 
\begin{itemize}
\item{Meta-Learning for \acrshort{DG} (\textbf{MLDG}~\cite{li2017learning})} is the first proposed meta-learning strategy that splits meta train/test and performs gradient update each minibatch, this makes the model trained on one domain to perform well on another domain.
\end{itemize}

\textit{\textbf{Domain-invariant learning}}: 
\begin{itemize}
\item{Invariant Risk Minimization (\textbf{IRM}~\cite{arjovsky2020irm})} learns invariant feature representation such that the optimal linear classifier on top of that representation matches across domains.

\item{Deep CORrelation ALignment (\textbf{CORAL}~\cite{sun2016corall})} matches the mean and covariance (second-order statistics) of features across training domain distributions at some level of representation.

\item{Maximum Mean Discrepancy (\textbf{MMD}~\cite{li2018mmd})} employs the adversarial technique and the maximum mean discrepancy (MMD~\cite{gretton12a}) criteria to align feature distribution across domain.

\item{Domain Adversarial Neural Networks (\textbf{DANN}~\cite{ganin2016dann})} uses an adversarial network to learn feature representation that matches across domains.

\item{Class-conditional DANN (\textbf{CDANN}~\cite{li2018cdann})} is a variant of DANN that matches the feature conditional distributions across domains, for all class labels to enable alignment of multimodal distributions.

\item{Risk Extrapolation (\textbf{VREx}~\cite{krueger2021vrex})} approximates IRM to reduce the variance of error averages across domains.

\end{itemize}

\textit{\textbf{Augmenting data}}: 
\begin{itemize}
\item{Inter-domain Mixup (\textbf{Mixup}~\cite{xu2019adversarial,yan2020improve,wang2020heterogeneous})} performs ERM on linear interpolations of examples from random pairs of domains and their labels.

\item{Style-Agnostic Networks (\textbf{SagNets}~\cite{nam2021sagnets})} promote representations of data that ignore image style and focus on content.

\item{Representation Self Challenging (\textbf{RSC}~\cite{huang2020rsc})} learns robust neural networks by iteratively dropping out (challenging) the most activated  (important) features.
\end{itemize}

\subsection{Implementation details}\label{apd:implementation}
In this appendix, we describe the data-processing techniques, neural network architectures, hyper-parameters, and details for reproducing our experiments.  We use similar settings from DomainBed~\cite{gulrajani2020domainbed} for a fair comparison. 

\textbf{Data processing techniques.}
For experiments related to the MNIST dataset, we receive an image with input size 28 × 28 x $d$ pixels (where $d$ is the image dimension, in which $d=2$ for Colored-MNIST, $d=1$ for Rotated-MNIST, and $d=3$ for Background-Colored-MNIST), and divide all the digits evenly among domains. For the remaining datasets, we augment training data using the following protocol: crops of random size and aspect ratio, resizing to 224 × 224 x 3 pixels, random horizontal flips, random color jitter, grayscaling the image with $10\%$ probability, and normalization using the ImageNet channel statistics.

\textbf{Architectures and hyper-parameters.}
We list the details of the backbone network, value of hyper-parameters used for each dataset in Table~\ref{tab:hyper-params}. We optimize all models using Adam~\cite{kingma2017adam} optimizer and employ the training-domain validation set technique for model selection. In particular, for all datasets, we first merge the raw training, validation, and test-set, then, we run the test three times with three different seeds. For each random seed, we randomly split training and validation from each source domain into 80\% and 20\% splits. and choose the model maximizing the accuracy on the validation set, then compute performance on the domain test-sets after 5000 iterations.\\
For experiments related to the MNIST dataset, we use MNIST-ConvNet backbone that have the structure following:\\
\textit{Conv2D (in=d, out=64) $\rightarrow$ Relu $\rightarrow$ GroupNorm (groups=8) $\rightarrow$ Conv2D (in=64, out=128, stride=2) $\rightarrow$ ReLU $\rightarrow$ GroupNorm (8 groups) $\rightarrow$ Conv2D (in=128, out=128) $\rightarrow$ ReLU $\rightarrow$ GroupNorm (8 groups) $\rightarrow$ Conv2D (in=128, out=128) $\rightarrow$ ReLU $\rightarrow$ GroupNorm (8 groups) $\rightarrow$ Global average-pooling.}\\
For the remaining datasets, we finetune the ResNet-50 model~\cite{resnet} which is pre-trained on ImageNet to avoid the inconsistent choice of network architecture in prior works. We customize the “ResNet-50” by replacing the final (softmax) layer and fine-tune the entire network. Since mini-batches from different domains follow different distributions, batch normalization degrades \acrshort{DG} algorithms~\cite{seo2019l}. Therefore, we freeze all batch normalization layers before fine-tuning.

\vskip -0.2in
\begin{table}[ht!]
\caption{Condition architectures, hyper-parameters, and their default values in our experiments.}
\label{tab:hyper-params}
\centering
\begin{tabular}{lll}
\toprule
\textbf{Condition} & \textbf{Hyper-parameters} & \textbf{Default value} \\
\midrule
\multirow{3}{*}{MNIST-ConvNet} & learning rate & 0.001\\ 
  & batch size & 64\\ 
  & generator learning rate & 0.001\\
  & discriminator learning rate & 0.001\\
\midrule
\multirow{4}{*}{ResNet} & learning rate & 0.00005\\ 
 & batch size & 32\\ 
 & batch size (if ARM) & 8\\  
 & generator learning rate & 0.00005\\
  & discriminator learning rate & 0.00005\\
\midrule
\multirow{7}{*}{DANN, C-DANN} & lambda & 1.0\\ 
  & discriminator weight decay & 0\\ 
  & discriminator steps & 1\\
  & discriminator width & 256\\
  & discriminator depth & 3\\
  & discriminator dropout & 0\\
  & discriminator grad penalty & 0\\
  & Adam $\beta_{1}$ & 0.5\\
\midrule
\multirow{1}{*}{DRO} & eta & 0.01\\
\midrule
\multirow{2}{*}{IRM} & labmda & 100\\
  & warmup iterations & 500\\
\midrule
\multirow{1}{*}{Mixup} & alpha & 0.2\\
\midrule
\multirow{1}{*}{MLDG} & beta & 1\\
\midrule
\multirow{1}{*}{MMD} & gamma & 1\\
\midrule
\multirow{1}{*}{MTL} & ema & 0.99\\
\midrule
\multirow{2}{*}{RSC} & feature drop percentage & 1/3\\
  & batch drop percentage & 1/3\\
\midrule
\multirow{1}{*}{SagNets} & adversary weight & 0.1\\
\midrule
\multirow{2}{*}{VREx} & lambda & 10\\
  & warmup iterations & 500\\
\midrule
\multirow{3}{*}{\acrshort{DSDI}} & domain-invariant weight & 1.0\\
  & domain-specific weight & 1.0\\
  & adversary weight & 1.0\\
\bottomrule
\end{tabular}
\end{table}

\textbf{Dataset, source code, and computing system.}
The source code is provided in the zip file, including scripts to download the dataset, setup for environment configuration, our provided code, and extending code from DomainBed~\cite{gulrajani2020domainbed} library (detail in README.md). We run the code on a single GPU: NVIDIA DGX-1 Tesla A100-SXM4-40GB with 12 CPUs: Intel(R) Core(TM) i7-8700 CPU @ 3.20GHz, RAM: 32GB, and require 40GB available disk space for storage.

\subsection{Empirical result details}\label{apd:results}
In this appendix, we show our full results and explain them in more detail when compared with other baseline methods in each benchmark dataset.

\vskip -0.2in
\begin{table}[ht!]
\caption{Classification accuracy (\%) on Colored-MNIST.}
\label{tab:c-mnist}
\centering
\begin{tabular}{lcccc}
\toprule
\textbf{Method} & \textbf{10\% flip} & \textbf{20\% flip} & \textbf{90\% flip} & \textbf{Average} \\
\midrule
ERM~\cite{vapnik1998erm} & 71.7$\pm$0.1 & 72.9$\pm$0.2 & 10.0$\pm$0.1 & 51.5\\
IRM~\cite{arjovsky2020irm} & 72.5$\pm$0.1 & 73.3$\pm$0.5 & 10.2$\pm$0.3 & 52.0\\
GroupDRO~\cite{sagawa2020dro} & 73.1$\pm$0.3 & 73.2$\pm$0.2 & 10.0$\pm$0.2 & 52.1\\
Mixup~\cite{xu2019adversarial,yan2020improve,wang2020heterogeneous} & 72.7$\pm$0.4 & 73.4$\pm$0.1 & 10.1$\pm$0.1 & 52.1\\
MLDG~\cite{li2017learning} & 71.5$\pm$0.2 & 73.1$\pm$0.2 & 9.8$\pm$0.1 & 51.5\\
CORAL~\cite{sun2016corall} & 71.6$\pm$0.3 & 73.1$\pm$0.1 & 9.9$\pm$0.1 & 51.5\\
MMD~\cite{li2018mmd} & 71.4$\pm$0.3 & 73.1$\pm$0.2 & 9.9$\pm$0.3 & 51.5\\
DANN~\cite{ganin2016dann} &  71.4$\pm$0.9 & 73.1$\pm$0.1 & 10.0$\pm$0.0 & 51.5\\
CDANN~\cite{li2018cdann} & 72.0$\pm$0.2 & 73.0$\pm$0.2 & 10.2$\pm$0.1 & 51.7\\
MTL~\cite{blanchard2011generalizing,blanchard2021mtl} & 70.9$\pm$0.2 & 72.8$\pm$0.3 &  \textbf{10.5$\pm$0.1} & 51.4\\
SagNets~\cite{nam2021sagnets} & 71.8$\pm$0.2 & 73.0$\pm$0.2 & 10.3$\pm$0.0 & 51.7\\
ARM~\cite{zhang2021arm} & \textbf{82.0$\pm$0.5} &  \textbf{76.5$\pm$0.3} & 10.2$\pm$0.0 & \textbf{56.2}\\
VREx~\cite{krueger2021vrex} & 72.4$\pm$0.3 & 72.9$\pm$0.4 & 10.2$\pm$0.0 & 51.8\\
RSC~\cite{huang2020rsc} & 71.9$\pm$0.3 & 73.1$\pm$0.2 & 10.0$\pm$0.2 & 51.7\\
\acrshort{DSDI} (Ours) & 73.4$\pm$0.2 & 73.1$\pm$0.3 & 10.1$\pm$0.2 & 52.2\\
\bottomrule
\end{tabular}
\end{table}

\textbf{Colored-MNIST.}
Table~\ref{tab:c-mnist} compares our results with the mentioned baseline on the Rotate-MNIST dataset. The average results show our model is not able to achieve a higher result than other methods, especially when compared to ARM~\cite{zhang2021arm} with 56.2\% on average while our result is only 52.2\%. However, it is worth noticing that if the unseen domain has a label-digit correlation which is reversed with source domains (i.e., the unseen domain is 90\% flip color), the performance of all models including ARM~\cite{zhang2021arm} and IRM~\cite{arjovsky2020irm} (the original paper proposed this dataset), drop significantly, only having around 10\% accuracy. This implies that not only our \acrshort{DSDI}, but also all models still concentrate on the color features in this challenging dataset. The color will hurt model performance in this dataset because it will be flipped randomly with a higher probability in the "90\% flip color" domain. Therefore, the model should not rely on color features and only focus on digits.

\vskip -0.2in
\begin{table}[ht!]
\caption{Classification accuracy (\%) on Rotated-MNIST.}
\label{tab:r-mnist}
\centering
\scalebox{0.9}{
\begin{tabular}{lccccccc}
\toprule
\textbf{Method} & \textbf{0} & \textbf{15} & \textbf{30} & \textbf{45} & \textbf{60} & \textbf{75} & \textbf{Average} \\
\midrule
ERM~\cite{vapnik1998erm} & 95.9$\pm$0.1 & 98.9$\pm$0.0 & 98.8$\pm$0.0 & 98.9$\pm$0.0 & 98.9$\pm$0.0 & 96.4$\pm$0.0 & 98.0\\
IRM~\cite{arjovsky2020irm} & 95.5$\pm$0.1 & 98.8$\pm$0.2 & 98.7$\pm$0.1 & 98.6$\pm$0.1 & 98.7$\pm$0.0 & 95.9$\pm$0.2 & 97.7\\
GroupDRO~\cite{sagawa2020dro} & 95.6$\pm$0.1 & 98.9$\pm$0.1 & 98.9$\pm$0.1 & 99.0$\pm$0.0 & 98.9$\pm$0.0 & \textbf{96.5$\pm$0.2} & 98.0\\
Mixup~\cite{xu2019adversarial,yan2020improve,wang2020heterogeneous} & 95.8$\pm$0.3 & 98.9$\pm$0.0 & 98.9$\pm$0.0 & 98.9$\pm$0.0 & 98.8$\pm$0.1 & 96.5$\pm$0.3 & 98.0\\
MLDG~\cite{li2017learning} & 95.8$\pm$0.1 & 98.9$\pm$0.1 & 99.0$\pm$0.0 & 98.9$\pm$0.1 & 99.0$\pm$0.0 & 95.8$\pm$0.3 & 97.9\\
CORAL~\cite{sun2016corall} & 95.8$\pm$0.3 & 98.8$\pm$0.0 & 98.9$\pm$0.0 & 99.0$\pm$0.0 & 98.9$\pm$0.1 & 96.4$\pm$0.2 & 98.0\\
MMD~\cite{li2018mmd} & 95.6$\pm$0.1 & 98.9$\pm$0.1 & 99.0$\pm$0.0 & 99.0$\pm$0.0 & 98.9$\pm$0.0 & 96.0$\pm$0.2 & 97.9\\
DANN~\cite{ganin2016dann} & 95.0$\pm$0.5 & 98.9$\pm$0.1 & 99.0$\pm$0.0 & 99.0$\pm$0.1 & 98.9$\pm$0.0 & 96.3$\pm$0.2 & 97.8\\
CDANN~\cite{li2018cdann} & 95.7$\pm$0.2 & 98.8$\pm$0.0 & 98.9$\pm$0.1 & 98.9$\pm$0.1 & 98.9$\pm$0.2 & 96.1$\pm$0.3 & 97.9\\
MTL~\cite{blanchard2011generalizing,blanchard2021mtl} & 95.6$\pm$0.1 & 99.0$\pm$0.1 & 99.0$\pm$0.0 & 98.9$\pm$0.1 & 99.0$\pm$0.1 & 95.8$\pm$0.2 & 97.9\\
SagNets~\cite{nam2021sagnets} & 95.9$\pm$0.3 & 98.9$\pm$0.1 & 99.0$\pm$0.1 & \textbf{99.1$\pm$0.0} & 99.0$\pm$0.1 & 96.3$\pm$0.1 & 98.0\\
ARM~\cite{zhang2021arm} & \textbf{96.7$\pm$0.2} & \textbf{99.1$\pm$0.0} & 99.0$\pm$0.0 & 99.0$\pm$0.1 & 99.1$\pm$0.1 & 96.5$\pm$0.4 & \textbf{98.2}\\
VREx~\cite{krueger2021vrex} & 95.9$\pm$0.2 & 99.0$\pm$0.1 & 98.9$\pm$0.1 & 98.9$\pm$0.1 & 98.7$\pm$0.1 & 96.2$\pm$0.2 & 97.9\\
RSC~\cite{huang2020rsc} & 94.8$\pm$0.5 & 98.7$\pm$0.1 & 98.8$\pm$0.1 & 98.8$\pm$0.0 & 98.9$\pm$0.1 & 95.9$\pm$0.2 & 97.9\\
\acrshort{DSDI} (Ours) & 96.0$\pm$0.1 & 98.8$\pm$0.1 & \textbf{99.1$\pm$0.0} & 98.9$\pm$0.0 & \textbf{99.2$\pm$0.1} & 96.2$\pm$0.1 & 98.0\\
\bottomrule
\end{tabular}}
\end{table}

\textbf{Rotated-MNIST.}
In Table~\ref{tab:r-mnist}, we compare our results with the mentioned baseline on the Rotated-MNIST dataset. It can be seen that this dataset only contains domain-invariant information while the domain-specific information is limited due to background-less images in the MNIST dataset. However, our \acrshort{DSDI} model still has a competitive result with $98\%$ on average when compared with other methods that concentrate on learning domain-invariant techniques such as IRM, Corral, MMD, DANN, CDANN, or VREx. This proves that our model still preserves domain-invariant information and the effectiveness of our adversarial training technique for extracting these features.

\vskip -0.2in
\begin{table}[ht!]
\caption{Classification accuracy (\%) on VLCS.}
\label{tab:vlcs}
\centering
\begin{tabular}{lccccc}
\toprule
\textbf{Method} & \textbf{C} & \textbf{L} & \textbf{S} & \textbf{V} & \textbf{Average} \\
\midrule
ERM~\cite{vapnik1998erm} & 97.7$\pm$0.4 & 64.3$\pm$0.9 & 73.4$\pm$0.5 & 74.6$\pm$1.3 & 77.5\\
IRM~\cite{arjovsky2020irm} & 98.6$\pm$0.1 & 64.9$\pm$0.9 & 73.4$\pm$0.6 & 77.3$\pm$0.9 & 78.5\\
GroupDRO~\cite{sagawa2020dro} & 97.3$\pm$0.3 & 63.4$\pm$0.9 & 69.5$\pm$0.8 & 76.7$\pm$0.7 & 76.7\\
Mixup~\cite{xu2019adversarial,yan2020improve,wang2020heterogeneous} & 98.3$\pm$0.6 & 64.8$\pm$1.0 & 72.1$\pm$0.5 & 74.3$\pm$0.8 & 77.4\\
MLDG~\cite{li2017learning} & 97.4$\pm$0.2 & 65.2$\pm$0.7 & 71.0$\pm$1.4 & 75.3$\pm$1.0 & 77.2\\
CORAL~\cite{sun2016corall} & 98.3$\pm$0.1 & 66.1$\pm$1.2 & 73.4$\pm$0.3 & 77.5$\pm$1.2 & 78.8\\
MMD~\cite{li2018mmd} & 97.7$\pm$0.1 & 64.0$\pm$1.1 & 72.8$\pm$0.2 & 75.3$\pm$3.3 & 77.5\\
DANN~\cite{ganin2016dann} & \textbf{99.0$\pm$0.3} & 65.1$\pm$1.4 & 73.1$\pm$0.3 & 77.2$\pm$0.6 & 78.6\\
CDANN~\cite{li2018cdann} & 97.1$\pm$0.3 & 65.1$\pm$1.2 & 70.7$\pm$0.8 & 77.1$\pm$1.5 & 77.5\\
MTL~\cite{blanchard2011generalizing,blanchard2021mtl} & 97.8$\pm$0.4 & 64.3$\pm$0.3 & 71.5$\pm$0.7 & 75.3$\pm$1.7 & 77.2\\
SagNets~\cite{nam2021sagnets} & 97.9$\pm$0.4 & 64.5$\pm$0.5 & 71.4$\pm$1.3 & 77.5$\pm$0.5 & 77.8\\
ARM~\cite{zhang2021arm} & 98.7$\pm$0.2 & 63.6$\pm$0.7 & 71.3$\pm$1.2 & 76.7$\pm$0.6 & 77.6\\
VREx~\cite{krueger2021vrex} & 98.4$\pm$0.3 & 64.4$\pm$1.4 & 74.1$\pm$0.4 & 76.2$\pm$1.3 & 78.3\\
RSC~\cite{huang2020rsc} & 97.9$\pm$0.1 & 62.5$\pm$0.7 & 72.3$\pm$1.2 & 75.6$\pm$0.8 & 77.1\\
\acrshort{DSDI} (Ours) & 97.6$\pm$0.1 & \textbf{66.4$\pm$0.4} & \textbf{74.0$\pm$0.6} & \textbf{77.8$\pm$0.7} & \textbf{79.0}\\
\bottomrule
\end{tabular}
\end{table}

\textbf{VLCS.}
In Table~\ref{tab:vlcs}, we compare our model's performance on VLCS dataset. It shows \acrshort{DSDI} archives the highest score on average with $79.0\%$, having competitive results on Caltech101 and dominating other baselines on three domains, including LabelMe, Sun09, and VOC2007. Interestingly, we observe that there are many samples in this dataset that miss the object and only contain a background such as an image of the bird but only having a sky picture or car image but only contain houses in the city. Therefore, the reason why our model outperforms other baselines could be explained by the fact that their domain-invariant method could not capture this domain-specific information (sky, houses), and so have a poor performance. Meanwhile, when comparing with other domain-specific based methods, the reason for our higher results would be that their method only concentrates on domain-specific techniques, and so, in some background-less images (e.g., only bird, car), these methods provide inferior domain-invariant information to our techniques. In contrast, due to considering disentangle domain-invariant and domain-specific features, our model captures both this useful information, hence, outperforms their results.

\vskip -0.2in
\begin{table}[ht!]
\caption{Classification accuracy (\%) on PACS.}
\label{tab:pacs-ResNet-18}
\centering
\begin{tabular}{lccccc}
\toprule
\textbf{Method} & \textbf{A} & \textbf{C} & \textbf{P} & \textbf{S} & \textbf{Average} \\
\midrule
ERM~\cite{vapnik1998erm} & 84.7$\pm$0.4 & \textbf{80.8$\pm$0.6} & 97.2$\pm$0.3 & 79.3$\pm$1.0 & 85.5\\
IRM~\cite{arjovsky2020irm} & 84.8$\pm$1.3 & 76.4$\pm$1.1 & 96.7$\pm$0.6 & 76.1$\pm$1.0 & 83.5\\
GroupDRO~\cite{sagawa2020dro} & 83.5$\pm$0.9 & 79.1$\pm$0.6 & 96.7$\pm$0.3 & 78.3$\pm$2.0 & 84.4\\
Mixup~\cite{xu2019adversarial,yan2020improve,wang2020heterogeneous} &86.1$\pm$0.5 & 78.9$\pm$0.8 & 97.6$\pm$0.1 & 75.8$\pm$1.8 & 84.6\\
MLDG~\cite{li2017learning} & 85.5$\pm$1.4 & 80.1$\pm$1.7 & 97.4$\pm$0.3 & 76.6$\pm$1.1 & 84.9\\
CORAL~\cite{sun2016corall} & \textbf{88.3$\pm$0.2} & 80.0$\pm$0.5 & 97.5$\pm$0.3 & 78.8$\pm$1.3 & 86.2\\
MMD~\cite{li2018mmd} & 86.1$\pm$1.4 & 79.4$\pm$0.9 & 96.6$\pm$0.2 & 76.5$\pm$0.5 & 84.6\\
DANN~\cite{ganin2016dann} & 86.4$\pm$0.8 & 77.4$\pm$0.8 & 97.3$\pm$0.4 & 73.5$\pm$2.3 & 83.6\\
CDANN~\cite{li2018cdann} &  84.6$\pm$1.8 & 75.5$\pm$0.9 & 96.8$\pm$0.3 & 73.5$\pm$0.6 & 82.6\\
MTL~\cite{blanchard2011generalizing,blanchard2021mtl} & 87.5$\pm$0.8 & 77.1$\pm$0.5 & 96.4$\pm$0.8 & 77.3$\pm$1.8 & 84.6\\
SagNets~\cite{nam2021sagnets} & 87.4$\pm$1.0 & 80.7$\pm$0.6 & 97.1$\pm$0.1 & \textbf{80.0$\pm$0.4} & \textbf{86.3}\\
ARM~\cite{zhang2021arm} & 86.8$\pm$0.6 & 76.8$\pm$0.5 & 97.4$\pm$0.3 & 79.3$\pm$1.2 & 85.1\\
VREx~\cite{krueger2021vrex} & 86.0$\pm$1.6 & 79.1$\pm$0.6 & 96.9$\pm$0.5 & 77.7$\pm$1.7 & 84.9\\
RSC~\cite{huang2020rsc} & 85.4$\pm$0.8 & 79.7$\pm$1.8 & 97.6$\pm$0.3 & 78.2$\pm$1.2 & 85.2\\
\acrshort{DSDI} (Ours) & 87.7$\pm$0.4 & 80.4$\pm$0.7 & \textbf{98.1$\pm$0.3} & 78.4$\pm$1.2 & 86.2\\
\bottomrule
\end{tabular}
\end{table}

\textbf{PACS.}
Table~\ref{tab:pacs-ResNet-18} compares our model with other methods on PACS dataset. It shows that if the target domain is either photo or art, which include more informative backgrounds such as dogs in the yard or guitars lying on a table, \acrshort{DSDI} achieved $98.1\%$ and $87.7\%$ respectively. It is higher than MLDG, which also uses the meta-training technique, but for all representations including domain-invariant. This reveals that although meta-training is essential in \acrshort{DG}, it is only for domain-specific features which need to be adapted to new domains.  Moreover, when comparing with other domain-specific based techniques such as GroupDRO, MTL, and ARM, the results showed that domain-specific features from our model are more helpful than theirs.\\
Meanwhile, due to still considering disentangled domain-invariant features, in two remaining domains of cartoon and sketch, which have a white background, our model still achieves $80.4 \%$ and $78.4\%$. It shows competitive results with other baselines that are based on domain-invariant learning such as DANN, C-DANN, CORAL, MMD, IRM, and VREx. It can be explained by the fact that we only consider meta-training on the domain-specific features, and so still retain informative domain-invariant. As a result, our \acrshort{DSDI} achieved a competitive accuracy with $86.2\%$ on average across test domains on PACS.

\vskip -0.2in
\begin{table}[ht!]
\caption{Classification accuracy (\%) on Office-Home.}
\label{tab:office-home}
\centering
\begin{tabular}{lccccc}
\toprule
\textbf{Method} & \textbf{A} & \textbf{C} & \textbf{P} & \textbf{R} & \textbf{Average} \\
\midrule
ERM~\cite{vapnik1998erm} & 61.3$\pm$0.7 & 52.4$\pm$0.3 & 75.8$\pm$0.1 & 76.6$\pm$0.3 & 66.5 \\
IRM~\cite{arjovsky2020irm} & 58.9$\pm$2.3 & 52.2$\pm$1.6 & 72.1$\pm$2.9 & 74.0$\pm$2.5 & 64.3\\
GroupDRO~\cite{sagawa2020dro} & 60.4$\pm$0.7 & 52.7$\pm$1.0 & 75.0$\pm$0.7 & 76.0$\pm$0.7 & 66.0\\
Mixup~\cite{xu2019adversarial,yan2020improve,wang2020heterogeneous} &62.4$\pm$0.8 & 54.8$\pm$0.6 & \textbf{76.9$\pm$0.3} & 78.3$\pm$0.2 & 68.1\\
MLDG~\cite{li2017learning} & 61.5$\pm$0.9 & 53.2$\pm$0.6 & 75.0$\pm$1.2 & 77.5$\pm$0.4 & 66.8\\
CORAL~\cite{sun2016corall} & 65.3$\pm$0.4 & 54.4$\pm$0.5 & 76.5$\pm$0.1 & 78.4$\pm$0.5 & 68.7\\
MMD~\cite{li2018mmd} & 60.4$\pm$0.2 & 53.3$\pm$0.3 & 74.3$\pm$0.1 & 77.4$\pm$0.6 & 66.3\\
DANN~\cite{ganin2016dann} & 59.9$\pm$1.3 & 53.0$\pm$0.3 & 73.6$\pm$0.7 & 76.9$\pm$0.5 & 65.9\\
CDANN~\cite{li2018cdann} &  61.5$\pm$1.4 & 50.4$\pm$2.4 & 74.4$\pm$0.9 & 76.6$\pm$0.8 & 65.8\\
MTL~\cite{blanchard2011generalizing,blanchard2021mtl} & 61.5$\pm$0.7 & 52.4$\pm$0.6 & 74.9$\pm$0.4 & 76.8$\pm$0.4 & 66.4\\
SagNets~\cite{nam2021sagnets} & 63.4$\pm$0.2 & \textbf{54.8$\pm$0.4} & 75.8$\pm$0.4 & 78.3$\pm$0.3 & 68.1\\
ARM~\cite{zhang2021arm} & 58.9$\pm$0.8 & 51.0$\pm$0.5 & 74.1$\pm$0.1 & 75.2$\pm$0.3 & 64.8\\
VREx~\cite{krueger2021vrex} & 60.7$\pm$0.9 & 53.0$\pm$0.9 & 75.3$\pm$0.1 & 76.6$\pm$0.5 & 66.4\\
RSC~\cite{huang2020rsc} & 60.7$\pm$1.4 & 51.4$\pm$0.3 & 74.8$\pm$1.1 & 75.1$\pm$1.3 & 65.5\\
\acrshort{DSDI} (Ours) & \textbf{68.1$\pm$0.3} & 52.1$\pm$0.4 & 76.0$\pm$0.2 & \textbf{80.4$\pm$0.2} & \textbf{69.2}\\
\bottomrule
\end{tabular}
\end{table}

\textbf{Office-Home.}
We have the same observation on the Office-Home dataset results in Table~\ref{tab:office-home}. Our \acrshort{DSDI} model outperforms other methods, achieving $69.2\%$ on average. Particularly, in the Art and Real-world domain, which contains more information in the background than other domains such as a bed in the room or bike parked on the street, our model reached $68.1\%$ and $80.4\%$ correspondingly, significantly higher than other methods. This means that our domain-invariant features not only support generalization better but also our domain-specific ones cover helpful information in special scenarios such as backgrounds and colors related to objects in the classification task.

\vskip -0.2in
\begin{table}[ht!]
\caption{Classification accuracy (\%) on Terra Incognita.}
\label{tab:terra}
\centering
\begin{tabular}{lccccc}
\toprule
\textbf{Method} & \textbf{L100} & \textbf{L38} & \textbf{L43} & \textbf{L46} & \textbf{Average} \\
\midrule
ERM~\cite{vapnik1998erm} & 49.8$\pm$4.4 & 42.1$\pm$1.4 & 56.9$\pm$1.8 & 35.7$\pm$3.9 & 46.1 \\
IRM~\cite{arjovsky2020irm} &  54.6$\pm$1.3 & 39.8$\pm$1.9 & 56.2$\pm$1.8 & 39.6$\pm$0.8 & 47.6\\
GroupDRO~\cite{sagawa2020dro} & 41.2$\pm$0.7 & 38.6$\pm$2.1 & 56.7$\pm$0.9 & 36.4$\pm$2.1 & 43.2\\
Mixup~\cite{xu2019adversarial,yan2020improve,wang2020heterogeneous} & \textbf{59.6$\pm$2.0} & 42.2$\pm$1.4 & 55.9$\pm$0.8 & 33.9$\pm$1.4 & 47.9\\
MLDG~\cite{li2017learning} & 54.2$\pm$3.0 & \textbf{44.3$\pm$1.1} & 55.6$\pm$0.3 & 36.9$\pm$2.2 & 47.7\\
CORAL~\cite{sun2016corall} & 51.6$\pm$2.4 & 42.2$\pm$1.0 & 57.0$\pm$1.0 & 39.8$\pm$2.9 & 47.6\\
MMD~\cite{li2018mmd} & 41.9$\pm$3.0 & 34.8$\pm$1.0 & 57.0$\pm$1.9 & 35.2$\pm$1.8 & 42.2\\
DANN~\cite{ganin2016dann} & 51.1$\pm$3.5 & 40.6$\pm$0.6 & 57.4$\pm$0.5 & 37.7$\pm$1.8 & 46.7\\
CDANN~\cite{li2018cdann} & 47.0$\pm$1.9 & 41.3$\pm$4.8 & 54.9$\pm$1.7 & 39.8$\pm$2.3 & 45.8\\
MTL~\cite{blanchard2011generalizing,blanchard2021mtl} & 49.3$\pm$1.2 & 39.6$\pm$6.3 & 55.6$\pm$1.1 & 37.8$\pm$0.8 & 45.6\\
SagNets~\cite{nam2021sagnets} &  53.0$\pm$2.9 & 43.0$\pm$2.5 & \textbf{57.9$\pm$0.6} & 40.4$\pm$1.3 & \textbf{48.6}\\
ARM~\cite{zhang2021arm} & 49.3$\pm$0.7 & 38.3$\pm$2.4 & 55.8$\pm$0.8 & 38.7$\pm$1.3 & 45.5\\
VREx~\cite{krueger2021vrex} &  48.2$\pm$4.3 & 41.7$\pm$1.3 & 56.8$\pm$0.8 & 38.7$\pm$3.1 & 46.4\\
RSC~\cite{huang2020rsc} & 50.2$\pm$2.2 & 39.2$\pm$1.4 & 56.3$\pm$1.4 & \textbf{40.8$\pm$0.6} & 46.6\\
\acrshort{DSDI} (Ours) & 53.2$\pm$3.0 & 43.3$\pm$1.0 & 56.7$\pm$0.5 & 39.2$\pm$1.3 & 48.1\\
\bottomrule
\end{tabular}
\end{table}

\textbf{Terra Incognita.}
Table~\ref{tab:terra} compares the classification accuracy of our \acrshort{DSDI} with other baselines on Terra Incognita dataset. It shows that our method provides the second-best result on the average accuracy with $48.1\%$, only lost to SagNet which achieves $48.6\&$. It is worth noticing that this dataset lacks domain-specific features to fully exploit the advantages of our algorithm. It contains photographs of wild animals taken by camera traps; camera trap locations are different across domains. Since these cameras are static, different animals still have the same background, or in other words, there is no correlation between background and animal label in this dataset. Hence, it may lack domain-specific features, making our model rely mainly on domain-invariant features of the object animal.

\vskip -0.2in
\begin{table}[ht!]
\caption{Classification accuracy  (\%) on DomainNet.}
\label{tab:domain-net}
\centering
\scalebox{0.85}{
\begin{tabular}{lccccccc}
\toprule
\textbf{Method} & \textbf{C} & \textbf{I} & \textbf{P} & \textbf{Q} & \textbf{R} & \textbf{S} & \textbf{Average} \\
\midrule
ERM~\cite{vapnik1998erm} & 58.1$\pm$0.3 & 18.8$\pm$0.3 & 46.7$\pm$0.3 & 12.2$\pm$0.4 & 59.6$\pm$0.1 & 49.8$\pm$0.4 & 40.9\\
IRM~\cite{arjovsky2020irm} & 48.5$\pm$2.8 & 15.0$\pm$1.5 & 38.3$\pm$4.3 & 10.9$\pm$0.5 & 48.2$\pm$5.2 & 42.3$\pm$3.1 & 33.9\\
GroupDRO~\cite{sagawa2020dro} & 47.2$\pm$0.5 & 17.5$\pm$0.4 & 33.8$\pm$0.5 & 9.3$\pm$0.3 & 51.6$\pm$0.4 & 40.1$\pm$0.6 & 33.3\\
Mixup~\cite{xu2019adversarial,yan2020improve,wang2020heterogeneous} &55.7$\pm$0.3 & 18.5$\pm$0.5 & 44.3$\pm$0.5 & 12.5$\pm$0.4 & 55.8$\pm$0.3 & 48.2$\pm$0.5 & 39.2\\
MLDG~\cite{li2017learning} & 59.1$\pm$0.2 & 19.1$\pm$0.3 & 45.8$\pm$0.7 & \textbf{13.4$\pm$0.3} & 59.6$\pm$0.2 & 50.2$\pm$0.4 & 41.2\\
CORAL~\cite{sun2016corall} & 59.2$\pm$0.1 & \textbf{19.7$\pm$0.2} & 46.6$\pm$0.3 & 13.4$\pm$0.4 & 59.8$\pm$0.2 & 50.1$\pm$0.6 & 41.5\\
MMD~\cite{li2018mmd} & 32.1$\pm$13.3 & 11.0$\pm$4.6 & 26.8$\pm$11.3 & 8.7$\pm$2.1 & 32.7$\pm$13.8 & 28.9$\pm$11.9 & 23.4\\
DANN~\cite{ganin2016dann} & 53.1$\pm$0.2 & 18.3$\pm$0.1 & 44.2$\pm$0.7 & 11.8$\pm$0.1 & 55.5$\pm$0.4 & 46.8$\pm$0.6 & 38.3\\
CDANN~\cite{li2018cdann} &  54.6$\pm$0.4 & 17.3$\pm$0.1 & 43.7$\pm$0.9 & 12.1$\pm$0.7 & 56.2$\pm$0.4 & 45.9$\pm$0.5 & 38.3\\
MTL~\cite{blanchard2011generalizing,blanchard2021mtl} & 57.9$\pm$0.5 & 18.5$\pm$0.4 & 46.0$\pm$0.1 & 12.5$\pm$0.1 & 59.5$\pm$0.3 & 49.2$\pm$0.1 & 40.6\\
SagNets~\cite{nam2021sagnets} & 57.7$\pm$0.3 & 19.0$\pm$0.2 & 45.3$\pm$0.3 & 12.7$\pm$0.5 & 58.1$\pm$0.5 & 48.8$\pm$0.2 & 40.3\\
ARM~\cite{zhang2021arm} & 49.7$\pm$0.3 & 16.3$\pm$0.5 & 40.9$\pm$1.1 & 9.4$\pm$0.1 & 53.4$\pm$0.4 & 43.5$\pm$0.4 & 35.5\\
VREx~\cite{krueger2021vrex} & 47.3$\pm$3.5 & 16.0$\pm$1.5 & 35.8$\pm$4.6 & 10.9$\pm$0.3 & 49.6$\pm$4.9 & 42.0$\pm$3.0 & 33.6\\
RSC~\cite{huang2020rsc} & 55.0$\pm$1.2 & 18.3$\pm$0.5 & 44.4$\pm$0.6 & 12.2$\pm$0.2 & 55.7$\pm$0.7 & 47.8$\pm$0.9 & 38.9\\
\acrshort{DSDI} (Ours) & \textbf{62.1$\pm$0.3} & 19.1$\pm$0.4 & \textbf{49.4$\pm$0.4} & 12.8$\pm$0.7 & \textbf{62.9$\pm$0.3} & \textbf{50.4$\pm$0.4} & \textbf{42.8}\\
\bottomrule
\end{tabular}}
\end{table}

\textbf{DomainNet.}
Finally, we experiment on DomainNet, a known large-scale dataset. In Table~\ref{tab:domain-net}, we observe that our \acrshort{DSDI} not only has the highest average number with $42.8\%$ but also dominates other methods in 4 domain tests. This implies that when the number of dataset increases, our method can extract more relevant information for complex tasks, such as classifying 345 classes in DomainNet. These results also mean that our model has a balance between informative domain-invariant and domain-specific features to adapt better to different environments than others, therefore showing the highest average in all settings. For instance, the "infograph" is a domain containing hundreds of words in the background (domain-specific relevant) to describe the object, even though \acrshort{DSDI} can not outperform the evaluation, it is still better than the many different methods. More importantly, in a sketch domain, which does not have a background link to the object, our model always achieves significantly better performance than other methods, and so implies that our domain-invariant is much better.

\subsection{Changing of Loss Functions}\label{apd:loss}
\begin{figure}[ht!]
\begin{center}
  \includegraphics[width=1.0\linewidth]{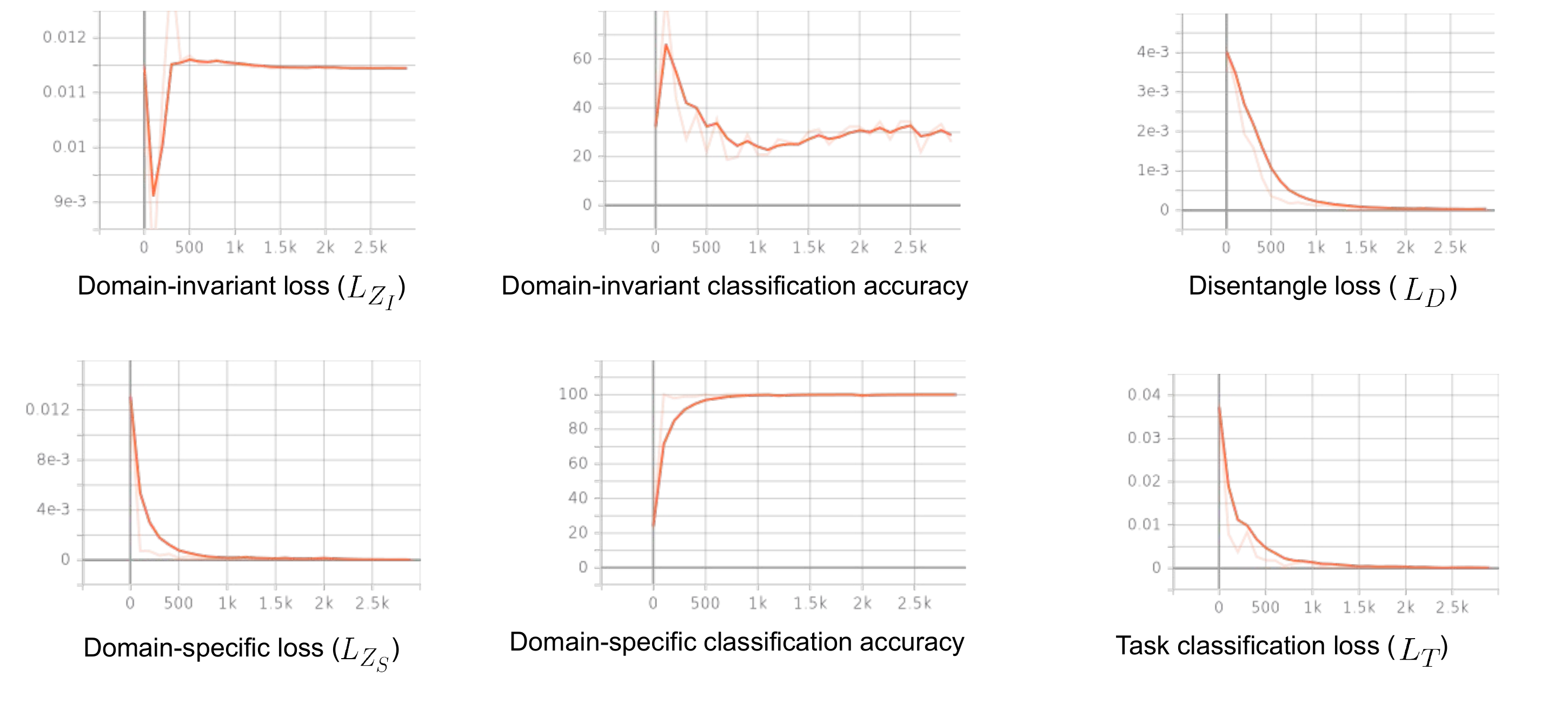}
\end{center}
  \caption{Losses visualization with tensorboard. Settings: PACS with ResNet-50, source domains include art, cartoon, sketch while target domain is photo.}
\label{fig:tensorboard}
\end{figure}

This appendix shows the behavior of \acrshort{DSDI}'s losses functions during the training time. Figure~\ref{fig:tensorboard} visualizes the changing of objective functions and their classification accuracy during training time. 

We observe that the classification accuracy of domain-invariant with respect to $L_{Z_{I}}$ loss decreases after a few interactions, then remains stable around $30\%$ during the training time. This means that the domain-invariant extractor $Q$ fools the domain-invariant discriminator $D_{I}$ successfully. Regarding domain-specific features, domain-specific extractor $R$ have extracted specific features by reducing $L_{Z_{S}}$ loss from domain-specific classifier $D_{S}$. Meanwhile, the disentangle loss $L_{D}$ decreases to zero, showing that the features extracted from $Q$ and $R$ are disentangled. Finally, the classifier loss $L_{T}$ is minimized close to zero in training samples, meaning that it preserves sufficient representation for domain-invariant and domain-specific in the classification task.

\subsection{Ablation study: Important of \acrshort{DSDI} on the benchmark dataset}\label{apd:ablation}
For further analysis in ablation study with PACS~\cite{li2017deeper}, a standard benchmark dataset in \acrshort{DG}. This appendix studies the performance efficiency of domain-specific and domain-invariant features per each target domain under different settings mentioned in the main paper. 

\begin{table}[ht!]
\caption{Classification accuracy (\%) on the benchmark dataset PACS. Ablation study shows impact of domain-invariant when combined with meta-training on domain-specific in our method.}
\label{tab:additional-results}
\centering
\begin{tabular}{lccccc}
\toprule
\textbf{Method} & \textbf{A} & \textbf{C} & \textbf{P} & \textbf{S} & \textbf{Average} \\
\midrule
DI & 84.5$\pm$0.6 & 77.4$\pm$0.8 & 98.0$\pm$0.3 & 74.7$\pm$1.2 & 83.7\\
DI-Meta & 84.4$\pm$0.5 & 77.3$\pm$0.7 & 97.5$\pm$0.4 & 75.1$\pm$1.4 & 83.6\\
DS & 85.2$\pm$0.5 & 77.9$\pm$0.9 & 98.0$\pm$0.3 & 72.5$\pm$1.3 & 83.4\\
DS-Meta & 87.1$\pm$0.4 & 79.2$\pm$0.7 & 98.2$\pm$0.3 & 74.1$\pm$1.2 & 84.7\\
DSDI-Without $L_D$ & 86.4$\pm$0.5 & 78.9$\pm$0.8 & 98.0$\pm$0.3 & 72.4$\pm$2.4 & 83.9\\
DSDI-Without Meta & 84.4$\pm$0.6 & 79.2$\pm$0.8 & 98.3$\pm$0.3 & 75.7$\pm$1.4 & 84.4\\
DSDI-Meta & 86.5$\pm$0.3 & 78.4$\pm$0.7 & \textbf{98.3$\pm$0.3} & 77.5$\pm$1.2 & 85.2\\
DSDI-Meta DI & 84.5$\pm$0.4 & 77.8$\pm$0.8 & 97.9$\pm$0.3 & 76.6$\pm$1.1 & 84.2\\
\acrshort{DSDI}-Meta DS (Ours) & \textbf{87.7$\pm$0.4} & \textbf{80.4$\pm$0.7} & 98.1$\pm$0.3 & \textbf{78.4$\pm$1.2} & \textbf{86.2}\\
\bottomrule
\end{tabular}
\end{table}

Table~\ref{tab:additional-results} shows that if the target domain is either photo, art, or cartoon, which has colors, the settings related to domain-specific features provide better performance than domain-invariant. For instance, meta-training on domain-specific (DS-Meta) reaches $87.1\%$, $79.2\%$ for the art and cartoon domain, respectively. In contrast, if the target domain is the sketch, which is without the color information, the settings related to domain-invariant outperform domain-specific features, such as $75.1\%$ for meta-training domain-invariant (DI-Meta). This is reasonable and confirms our assumptions that the domain-specific is color and background information while domain-invariant is the object's sketch.

Besides confirming the hypothesis of domain-specific and domain-invariant, we also observe that if we can learn both these features in the right strategy, the performance is even better. Specifically, when comparing between DSDI-Meta DI, which uses meta-training on domain-invariant, and \acrshort{DSDI}-Meta DS, which uses meta-training on domain-specific, \acrshort{DSDI}-Meta DS always achieves better performance in all target domain settings. As a result, achieved the highest on average with $86.2\%$ accuracy. It confirms the argument that, if we assume domain-invariant is stable across domains, the model should not optimize on these features. Instead, it should be used for domain-specific, which needs to be adapted on unseen domains. Meta-training on domain-invariant might lead to the model forgetting about the common because that procedure attempts to discover other independent contexts.

Finally, it is also worth noticing that without the disentanglement loss $L_D$, the average classification accuracy on target domains only achieves $83.9\%$, lower than around $2.5\%$ when compared with the highest one \acrshort{DSDI}-Meta DS which minimizing $L_D$. This shows that the disentangle method based on minimizing the covariance matrix in our framework work well and is essential to make domain-invariant and domain-specific independence, leading to an effective end-to-end framework to boost the generalization ability. 
\end{document}